\documentclass[preprint,12pt]{elsarticle}

 \usepackage{setspace}
\usepackage{float}
\usepackage{amsmath, amssymb, dsfont, mathrsfs, amsthm}
\usepackage[caption=false]{subfig}
\usepackage{enumitem}
\usepackage[ruled]{algorithm}
\usepackage{algpseudocode}
\usepackage{algcompatible}
\algrenewcommand\algorithmicindent{2em}
\algrenewcommand{\COMMENT}[1]{\hfill$\rhd$\,\text{#1}}
\usepackage{empheq} 
\usepackage{colonequals}
\usepackage{booktabs}
\usepackage{xcolor}
\usepackage[normalem]{ulem}
\usepackage{mathabx}
\usepackage{pifont}
\usepackage{multirow}
\usepackage{tablefootnote}
\newcommand{\Ding}[1]{\raisebox{-1pt}{\scalebox{1.1}{\ding{#1}}}}
\newcommand{\dingg}[1]{\raisebox{-0.5pt}{\scalebox{1.1}{\ding{#1}}}}
\makeatletter
\usepackage{tikz}
\usetikzlibrary{calc}
\newcommand*\circled[1]{\tikz[baseline=(char.base)]{
    \node[shape=circle, draw, inner sep=0pt, 
        minimum height={\f@size*1},] (char) {\vphantom{WAH1g}#1};}}

\usepackage{lineno}

\usepackage[margin=1.15in]{geometry}

\usepackage{natbib} 
\biboptions{authoryear,round}
 \usepackage[
                  breaklinks = true,
                 colorlinks = true,
                 linkcolor = red,
                 urlcolor  = black, 
                 citecolor = blue,
                 anchorcolor = green,
                 ]{hyperref}

\newcommand{\bX}{\mathbf{X}}

\newcommand{\bsP}{\boldsymbol{P}}

\newcommand{\cX}{\mathcal{X}}
\newcommand{\cE}{\mathcal{E}}

\newcommand{\cT}{\mathcal{T}}

\newcommand{\cH}{\mathcal{H}}

\newcommand{\R}{\mathbb{R}}

\newcommand{\N}{\mathbb{N}}

\newcommand{\sD}{\mathscr{D}}

\DeclareMathSymbol{\sm}{\mathbin}{AMSa}{"39}

\DeclareMathOperator{\val}{val}

\usepackage{soul}

\journal{Neurocomputing}

\begin{document}

\begin{frontmatter}

\title{Federated Dynamic Modeling and Learning \\ for Spatiotemporal Data Forecasting}      
\author[GRETTIA]{Thien~Pham} \ead{nhat-thien.pham@univ-eiffel.fr}
\author[ENTPE]{Angelo~Furno} \ead{angelo.furno@univ-eiffel.fr}
\author[SystemX]{Fa\"icel~Chamroukhi\corref{cor1}} \ead{faicel.chamroukhi@irt-systemx.fr}
\author[GRETTIA]{Latifa~Oukhellou} \ead{latifa.oukhellou@univ-eiffel.fr}
\cortext[cor1]{Corresponding author}%

\address[GRETTIA]{COSYS-GRETTIA, Gustave Eiffel University, 77420 France}
\address[ENTPE]{ENTPE, University of Lyon, and the LICIT-ECO7 University Gustave Eiffel, France }
\address[SystemX]{IRT-SystemX, 2 Boulevard Thomas Gobert, 91120 Palaiseau, France }

\begin{abstract}
This paper presents an advanced Federated Learning (FL) framework for forecasting complex spatiotemporal data, improving upon recent state-of-the-art models. In the proposed approach, the original Gated Recurrent Unit (GRU) module within previous Dynamic Spatial--Temporal Graph Convolutional Recurrent Network (DSTGCRN) modeling is first replaced with a Long Short-Term Memory (LSTM) network, enabling the resulting model to more effectively capture long-term dependencies inherent to time series data. The resulting architecture significantly improves the model's capacity to handle complex temporal patterns in diverse forecasting applications. Furthermore, the proposed FL framework integrates a novel Client-Side Validation (CSV) mechanism, introducing a critical validation step at the client level before incorporating aggregated parameters from the central server into local models, ensuring only the most effective updates are retained and improving both the robustness and accuracy of the forecasting model across clients. 
The efficiency of our approach is demonstrated through extensive experiments on real-world applications, including public datasets for multimodal transport demand forecasting and private datasets for Origin-Destination (OD) matrix forecasting in urban areas.
The results demonstrate substantial improvements over conventional methods, highlighting the framework's ability to capture complex spatiotemporal dependencies while preserving data privacy. This work not only provides a scalable and privacy-preserving solution for real-time, region-specific forecasting and management but also underscores the potential of leveraging distributed data sources in a FL context. 
We provide our algorithms as open-source on GitHub\footnote{\url{https://github.com/nhat-thien/Federated-LSTM-DSTGCRN}}.
\end{abstract}

%


\end{frontmatter}



\section{Introduction}
\label{sec:introduction}

Forecasting spatiotemporal data has become a critical area of research due to the increasing complexity and inter-connectivity of modern systems, such as urban transportation, energy grids, and environmental monitoring. Accurate predictions are essential for efficient planning, resource allocation, and minimizing disruptions. As urban areas and other complex systems continue to grow, the challenges of forecasting spatiotemporal data are exacerbated by the dynamic and heterogeneous nature of these systems.

Numerous models, ranging from traditional statistical methods to sophisticated machine learning techniques, have been employed to capture the intricate  spatiotemporal dependencies present in spatiotemporal data.
Simple models, such as autoregressive integrated moving average (ARIMA), exponential smoothing, and linear regression, have long been used in forecasting but often struggle to account for the complexities of spatiotemporal  interactions. These models typically assume stationarity and linearity, which can limit their effectiveness in dynamic and non-linear environments.

In recent years, the field has seen significant advancements with the introduction of machine learning and deep learning (DL) techniques. These methods have demonstrated remarkable success in capturing complex patterns in spatiotemporal data, offering significant improvements over traditional approaches.
Advanced models, including Graph Convolutional Networks (GCN) and Recurrent Neural Networks (RNN), have been increasingly utilized due to their ability to more effectively model spatial relationships and temporal dynamics. 
Long Short-Term Memory (LSTM), Gated Recurrent Unit (GRU), Convolutional Neural Networks (CNN) tailored for spatiotemporal prediction, and hybrid models that combine them, have demonstrated significant improvements in forecasting accuracy by leveraging the strengths of deep learning architectures. 
These models are particularly well-suited for capturing the non-linear and complex interactions inherent in spatiotemporal systems.

Graph Convolutional Recurrent Network (GCRN) models have established themselves as one of the most effective approaches for capturing spatial and temporal correlations in spatiotemporal data. By leveraging the strengths of GCNs and RNNs, GCRNs are particularly well-suited for modeling the dynamic interactions that characterize complex systems. Recent advancements in GCRN models have led to several variants tailored to specific aspects of spatiotemporal forecasting.

The Dynamic Spatial-Temporal Graph Convolutional Recurrent Network (DSTGCRN), introduced by \cite{GONG2024142581}, represents a state-of-the-art method for modeling complex spatiotemporal dependencies in multisource time series data. Initially developed for forecasting carbon emissions, the DSTGCRN's adaptable architecture makes it particularly well-suited for spatiotemporal prediction tasks. 
The model excels at capturing both spatial relationships and temporal dynamics, which are crucial for accurately predicting patterns that vary over time and across different locations. In this paper, we enhance the local forecasting model by integrating an LSTM network with the DSTGCRN, referred to as the LSTM-DSTGCRN model subsequently. 
This modification strengthens the model's ability to capture long-term temporal dependencies, which commonly occur in spatiotemporal data, thereby improving its effectiveness in the context of dynamic and evolving systems.

Despite advancements in forecasting, a significant challenge remains: many existing approaches rely on centralized datasets. In numerous cases, however, data are inherently distributed across various locations, making centralized access impractical. 
Additionally, models that depend on centralized data face several challenges, including privacy concerns, data ownership issues, and the logistical complexities of managing large-scale data repositories. 
Moreover, each organization's data may not capture all relevant characteristics and patterns, making it essential to collaborate and federate together to improve forecasts and predictions. 
However, directly sharing these datasets poses significant privacy risks and logistical hurdles, rendering collaborative efforts impractical. These challenges highlight the need for alternative methodologies that address both the practical and ethical limitations of centralized approaches.

Federated Learning (FL) \citep{McMahan2016FL, Konecn2016FederatedLS} presents an effective solution by enabling multiple clients to collaboratively train a model without sharing their raw data. 
Instead, only model parameters are exchanged and aggregated, thus preserving data privacy while leveraging the advantages of collective learning. This collaborative approach not only enhances the robustness of predictions but also maximizes the benefits derived from diverse data sources.

In this paper, we propose an enhanced FL framework that builds upon the LSTM-DSTGCRN model by incorporating a LSTM network in place of the original GRU. This modification enables the model to better capture long-term temporal dependencies, which are critical in various spatiotemporal forecasting tasks.

To further improve the FL process, we propose a novel Client-Side Validation (CSV) mechanism designed to enhance both model robustness and accuracy.
Specifically, after each round of server aggregation, clients validate the aggregated parameters before updating their local models. For each module, clients temporarily replace local parameters with the aggregated ones and compute the validation loss. 
If the loss improves, the new parameters are kept; otherwise, the original ones are restored. This selective update ensures that only beneficial changes are applied, improving overall model performance. 
Extensive experiments across various spatiotemporal datasets demonstrate that our approach achieves faster convergence and superior accuracy compared to state-of-the-art methods. This framework is particularly well-suited for tasks requiring privacy-preserving, collaborative learning across heterogeneous data sources.

The main contributions of this work are as follows: 
\begin{itemize}
    \item We enhance the DSTGCRN model by integrating LSTM networks instead of the original GRU. This modification significantly improves the models ability to capture long-term temporal dependencies, making it more effective for spatiotemporal forecasting tasks.
    
    \item We   design a FL framework 
     based on clients-server interaction, ensuring that 
     the prediction performance for all clients, based on the resulting collaboratively constructed model, is guaranteed to improve over the locally trained models.
    
    \item We propose a novel Client-Side Validation (CSV) mechanism that  controls the quality of aggregated  model parameters at the client level, ensuring that only beneficial updates are incorporated into local models. This mechanism enhances model robustness and accelerates convergence.

    \item We perform extensive experiments on real-world datasets, in two scenarios of application (multimodal transport demand forecasting and OD matrix forecasting), showcasing the superiority of our approach compared to traditional local models and state-of-the-art FL methods across various spatiotemporal forecasting tasks. 
\end{itemize}

The rest of the paper is organized as follows: Section~\ref{sec:related work} reviews related work in spatiotemporal forecasting and FL, Section~\ref{Problem formulation} defines the notations, formulates the problems of forecasting and FL with spatiotemporal data. Section~\ref{sec:LSTM-DSTGCRN} provides a detailed description of the LSTM-DSTGCRN model. In Section~\ref{sec: The proposed federating schemes}, we outline our FL framework and CSV process. Section~\ref{sec:experiments} presents our experiments for multimodal transport demand and OD matrix forecasting problems
Finally, Section~\ref{sec:results discussion} discusses the results and implications, and Section~\ref{sec:conclusion} concludes the paper with potential future research directions.

\section{Related Work}
\label{sec:related work}

Spatiotemporal data forecasting in transportation has seen significant advancements over the years, driven by the increasing availability of transport data and the development of sophisticated modeling techniques.
In recent years, deep learning models have emerged as powerful tools for  forecasting tasks due to their ability to learn complex patterns from large datasets. 
Recurrent neural networks (RNNs), particularly long short-term memory (LSTM) networks and gated recurrent units (GRUs), have been employed to capture temporal dependencies in transport data \citep{ShiLSTM2023, XuRNNTaxiIEEE2017, ShuFlowIEEE2022}.
However, these models primarily focus on modeling temporal patterns while often ignoring spatial dependencies, which are critical for understanding the interactions between different nodes in the  network. 
This limitation reduces their effectiveness in capturing the full complexity of spatiotemporal data, where both temporal and spatial relationships play a vital role in accurate forecasting.

Graph Convolutional Network (GCN) \citep{kipf2017GCN, Zhang2019} is an advanced approach, introduced to model spatial dependencies in graph-structured data by leveraging the inherent relationships between nodes and edges. Although GCNs are not specifically designed for transportation data, their ability to effectively represent networks as graphs, where nodes correspond to locations and edges capture connections or interactions, makes them well suited for analyzing transport networks.
Since its introduction, the combination of GCNs and RNNs has led to many advanced developments of spatiotemporal models. These hybrid architectures, known as Graph Convolutional Recurrent Networks (GCRNs), excel at simultaneously capturing spatial and temporal dependencies, making them powerful tools for dynamic network modeling and prediction tasks.

In \cite{YuYinZhu2017}, the authors introduced a spatio-temporal graph convolutional neural network that laid the groundwork for GCRNs frameworks in traffic forecasting.
Building on this foundation, \cite{Guo_Lin_Feng_Song_Wan_2019} proposed an attention-based  spatiotemporal  GCN to enhance the model's ability to focus on relevant spatial and temporal features, \cite{LiangHangJunlingAoli2019} explored temporal graph convolutional networks that account for external factors in traffic speed prediction.
Then, the works of \cite{Zheng2019GMANAG} with GMAN, \cite{Song2020} with STSGCN, and \cite{ChenLaiJin2020} with DST-GCNN models incorporate advanced attention mechanisms and GCNs to better capture the dynamic interactions and dependencies in both spatial and temporal dimensions.
However, these methods are limited to capturing common patterns across all traffic series and continue to depend on a pre-defined spatial connection graph.
Other extensions in this line of development include the LSGCN model proposed by \citep{ijcai2020p326}, which is designed for long short-term traffic prediction, and the OGCRNN model introduced by \citep{GuoGraphIEEE2021}, which integrates GCN and GRU to model traffic flow data. These models further demonstrate the effectiveness of GCRNs in capturing spatiotemporal dependencies in traffic systems.
Recent works have advanced GCRNs by incorporating more sophisticated modules to enhance their capabilities. For instance, \cite{HuLinWang2022} proposed a model that adaptively fuses geographical proximity and spatial heterogeneity information at each time step, further exploring the potential of GCRNs in traffic forecasting with pre-defined graph structures.
Another notable contribution is the Temporal Metrics-Based Aggregated Graph Convolution Network (TMAGCN) introduced by \cite{CHEN2023126662}. By incorporating temporal metrics, TMAGCN addresses the limitations of traditional distance-based metrics, offering a more accurate representation of real-world traffic conditions.


To address the limitation of the models based on pre-defined graph, in \cite{BAIYao2020}, the authors introduced the Adaptive Graph Convolutional Recurrent Network (AGCRN) model, designed to capture node-specific patterns to capture fine-grained spatial and temporal correlations
in traffic series automatically, making it highly flexible for real-world applications.
Building on this foundation, Spatiotemporal Adaptive Gated Graph Convolution Network proposed by \cite{BinXiaoying2020} enhances the ability to model complex spatiotemporal dependencies.

Another significant advancement is the Dynamic Spatial--Temporal Graph Recurrent Neural Network (DSTGRNN) proposed by \cite{XIA2024122381}. This framework combines a dynamic graph generator with a dynamic graph recurrent neural network and a novel fusion mechanism, enabling it to capture complex dynamic   spatiotemporal dependencies while integrating both static and dynamic graph features.
One of the most notable recent developments is the DS-STGCN model by \cite{HU2024323}, which further enhances traffic flow prediction by integrating node feature graphs, topology graphs, and time-slot feature graphs. This comprehensive approach effectively captures complex spatiotemporal dependencies, significantly improving prediction accuracy.
Other contribution in this line include the work of \cite{JingweiKarine2023}, which addresses the issue of missing data--a common challenge in traffic datasets caused by sensor malfunctions or communication errors.
Together, these advancements highlight the versatility and effectiveness of GCRN-based models in addressing the challenges of traffic forecasting and advancing the field.


The Dynamic Spatial--Temporal Graph Convolutional Recurrent Network (DSTGCRN) \cite{GONG2024142581} further extends these ideas by incorporating multihead attention mechanisms and adaptive graph convolutions, enabling the model to dynamically adjust to changing patterns in the data.
This makes the DSTGCRN particularly well-suited for multimodal transport demand forecasting, where transport modes and demand patterns can vary significantly over time and space.

FL has recently gained attention as a promising approach to training machine learning models across distributed datasets without requiring raw data sharing, thus preserving privacy and data security \cite{McMahan2016FL, Konecn2016FederatedLS}. 
In transport demand forecasting, FL allows multiple stakeholders, such as cities or transport agencies, to collaboratively train a global model while keeping their data decentralized. 
Various federating schemes have been proposed, including the original FedAvg \cite{McMahan2016FL}, which aggregates model parameters from local clients to form a global model. However, these approaches often assume homogeneous data distributions and may not effectively handle the heterogeneity, particularly, in multimodal transport data. {Recent FL methods address various aspects of the heterogeneity problem. Notably, FedDyn \citep{acar2021federatedlearningbaseddynamic}, FedProx \citep{li2020federatedoptimizationheterogeneousnetworks}, and FedOpt \citep{reddi2021adaptivefederatedoptimization} modify the update rule or add regularization to stabilize training across non-IID clients, while MOON \citep{li2021modelcontrastivefederatedlearning} targets client-specific feature gaps via contrastive learning for representation alignment and personalization. FedCurv \citep{casella2023benchmarkingfedavgfedcurvimage} constrains local drifts via curvature penalties through curvature-aware regularization.}

In the transportation domain, FL has been successfully applied across various aspects to improve prediction accuracy and operational efficiency. For instance, \cite{9082655} developed a privacy-preserving traffic flow prediction framework using FL, demonstrating its effectiveness in enhancing prediction models while safeguarding sensitive data. 
Similarly, \cite{Zeng2021MultiTaskFL} explored multi-task FL for traffic prediction, highlighting its application in route planning to optimize travel times. \cite{9798018} proposed FedTSE, a low-cost FL framework for privacy-preserved traffic state estimation in the Internet of Vehicles (IoV), emphasizing the practicality of FL in real-world applications.
Additionally, \cite{9684398} presented a network traffic prediction model that incorporates road traffic parameters, illustrating the potential of AI methods in vehicular ad-hoc networks (VANETs). Their subsequent work \cite{Sepasgozar2022} introduced Fed-NTP, a FL algorithm tailored for network traffic prediction in VANETs, further demonstrating FL's adaptability to various transportation contexts.

Moreover, recent advances by \cite{9737410} in graph representation-driven FL for urban traffic flow prediction underline the method's potential for edge computing applications in smart cities. A comprehensive review of FL applications in intelligent transportation systems can be found in \cite{Zhang2023FederatedLI} where the authors outlined recent developments and identified open challenges for future research.
%
Overall, these studies collectively highlight the transformative potential of FL in the transportation sector, effectively addressing the dual challenges of enhancing forecasting accuracy and ensuring data privacy in collaborative environments.

\section{Problem formulation}\label{Problem formulation}

\subsection{Spatiotemporal data forecasting}\label{subsec: Spatiotemporal data forecasting}

Let $M$ be the number of forecasting tasks we wish to address. For multimodal transport demand forecasting problem, these tasks could include, for example, forecasting the taxi demands of company \Ding{172}, the bike-sharing demand of company \Ding{173}, the bus demand of company \Ding{174}, etc. We refer to each entity that manages a specific forecasting task as a client. Each client has their network for service, e.g., pickup and drop-off locations (for taxis) or stations (for public transport), which we refer to as nodes.
For OD matrix forecasting problems, these tasks could include, for example, forecasting the flow of trips between zones in the city \circled{a}, the flow of trips between zones in the city \circled{b}, etc.
In this case, we also refer to each entity that own the data as a client, and the pairs of zones as nodes.

For $m\in\{1,\ldots,M\}$, let $N_m$ be the number of nodes in client $m$'s data. We denote by $X_{mn}(t)$, $n\in\{1,\ldots,N_m\}$, the value (i.e., the demand or the flow) at node $n$ of client $m$ at a time point $t\in\cT$. 
Here, we consider $\cT$ as a set of discrete increasing time points (e.g., hourly), denoted by $\cT = \{\ldots, t_{-1}, t_0, t_1,\ldots\}$.
We can see that for each $m\in\{1,,\ldots,M\}$ and $n\in\{1,\ldots,N_m\}$, $X_{mn}(\cdot)$ is a time series, i.e., a random function, with values in $\mathbb N$. 
Hence, our problem consists of forecasting such time series for all nodes of all clients given their historical data.

Furthermore, as will be described in the subsequent subsection, in our problem setting, we are subjected to respect the privacy constraints between the clients. In other words, while each client aims to develop a model that delivers the most accurate predictions for future outcomes across their entire network, they must do so without sharing raw data or disclosing any network details, such as the number of nodes or the scale of their data.

We denote by $\sD_m=\{(X_{m1}, Y_{m1}), \ldots, (X_{mN_m}, Y_{mN_m})\}$ the dataset that client $m$ owns. Here, for each $n\in\{1,\ldots,N_m\}$, 
$X_{mn}=\{X_{mn}(t),\ t\in\cT_{mn}\}$ is an observed time series (e.g., transport demand or OD flows), and
$Y_{mn}=\{Y_{mn}(t),\ t\in\cT_{mn}\}$ denotes the vector-valued time series of exogenous data, for example, the temperature, precipitation, hour of day, day of week, etc.
The notation $\cT_{mn}$ denotes the set of discrete time points that client $m$ observed their data at node $n$.  
For convenience, we assume $\cT_{mn}\equiv\cT=\{t_1, \ldots, t_T\}$ for all $m$ and $n$, with $t_1 < \ldots < t_T$ and $T$ being the historical length of the data.
Hence, our goal is to learn a forecasting model $\mathcal F_m$ for each client $m$, from their historical data $\sD_m$, to forecast the future demands (or flows) across all nodes in their network. 

Mathematically, at any time point $t_i$, given $p\in\N$ recent observations at all of $N_m$ nodes, 
client $m$ forecasts the values in their network for the $q\in\N$ upcoming time points as
\begin{align}\label{eq: general model}
    \bigg\{&X_{mn}(t), 
        \begin{array}{l}
        t\in \{t_{i+1}, \ldots, t_{i+q}\}, \\
        n\in\{1,\ldots,N_m\}
        \end{array} \bigg\}
    = \mathcal F_m \bigg(
    \bigg\{(X_{mn}(t), Y_{mn}(t)), 
        \begin{array}{l}
        t\in \{t_{i-p+1}, \ldots, t_{i}\}, \\
        n\in\{1,\ldots,N_m\}
        \end{array} \bigg\}; 
        \bsP_m\bigg),
\end{align}
where $\bsP_m$ is the parameter to be learned for the forecasting model $\mathcal F_m$.

The model $\mathcal{F}_m$ can take various forms depending on the complexity of the dynamics and the sophistication of the forecasting strategy employed.
If $\mathcal{F}_m$ is, e.g., a linear regression model, $\bsP_m$ might consist of regression coefficients that capture the linear relationships between past and future values. 
In this case, the model assumes that future values are a weighted sum of past values with some additional bias terms, making it straightforward but potentially limited in capturing complex patterns.

On the other hand, if $\mathcal{F}_m$ is a more advanced model such as a neural network or a graph-based model, $\bsP_m$ could include the weights and biases of a deep learning architecture, possibly comprising multiple layers and activation functions. 
For instance, in a Recurrent Neural Network (RNN) or a Graph Convolutional Network (GCN), $\bsP_m$ would encapsulate the intricate dependencies between nodes over time, enabling the model to capture non-linear and spatiotemporal correlations within the data.

The learning of $\bsP_m$ is driven by minimizing a suitable loss function, typically a function of the difference between the predicted and actual values. By training $\mathcal{F}_m$ on their respective datasets, each client aims to improve the accuracy and reliability of their forecasts, which are critical for optimizing resource allocation, scheduling, and overall network efficiency. 


In this paper, we employ the enhanced LSTM-DSTGCRN model, which builds upon the strengths of the original DSTGCRN model \citep{GONG2024142581}. 
By replacing the Gated Recurrent Unit (GRU) module with a Long Short-Term Memory (LSTM) network, the LSTM-DSTGCRN improves the ability to capture long-term temporal dependencies, particularly critical in transport demand forecasting. 
Unlike carbon emissions forecasting (the original application of the DSTGCRN model), where temporal patterns tend to evolve more gradually, transport demand often exhibits highly dynamic and recurring long-term trends, such as daily commuting patterns, seasonal variations, and event-driven surges. 
The integration of LSTM enables the model to better handle these complex temporal dynamics, making it more effective for transportation applications. 
So, in our case, each $\mathcal{F}_m$ in \eqref{eq: general model} is an LSTM-DSTGCRN model specifically configured for client $m$, and $\bsP_m$ contains multiple layers of three modules, each playing a vital role in capturing the spatiotemporal patterns in their respective forecasting tasks. The LSTM-DSTGCRN model and its parameter training are detailed in Section \ref{sec:LSTM-DSTGCRN}.

\subsection{Federated learning in spatiotemporal data forecasting} 
\label{subsec: Federated learning in spatiotemporal data forecasting}

In real-world scenarios, spatiotemporal data is often distributed across multiple entities or clients, each responsible for managing specific tasks within a designated geographic area or domain. 
For example, in transportation, clients may oversee different transport modes (e.g., taxis, buses, or bike-sharing systems) or operate in distinct regions. 
Similarly, in other domains such as energy consumption or environmental monitoring, clients may manage data from different sensors, zones, or systems.

This decentralized setup inherently limits each client's dataset to the specific characteristics of their local environment, leading to fragmented and incomplete data. Consequently, the data collected by any single client often lacks the comprehensive coverage needed to fully capture the diverse range of patterns, seasonal variations, and dynamic interactions present in spatiotemporal systems. 
This fragmentation creates a significant barrier to developing robust and generalizable forecasting models, as the insights derived from one client's data may not adequately represent the broader system dynamics.

Sharing datasets between clients could significantly enhance forecasting performance by providing access to a broader range of insights and patterns across diverse datasets. For instance, in transportation, combining data from multiple transport modes or regions could lead to more accurate and resilient models. 
Similarly, in OD matrix forecasting, collaboration between entities managing different zones could improve the prediction of trip flows across the entire network.

However, privacy concerns and data ownership restrictions add complexity to this scenario. Many organizations are unwilling or unable to share raw data due to the sensitive nature of the information, which may include proprietary details, personally identifiable data, or other confidential attributes. 


Therefore, it is crucial to develop a method that can leverage the collective knowledge of all clients without requiring them to share their raw data. 
By doing so, we can overcome the limitations of individual datasets and create a more holistic and accurate forecasting model. This is where FL comes into play. 
FL allows clients to collaboratively train a global model that captures a wider range of spatiotemporal patterns, improving the overall predictive performance for each client. 
This is particularly valuable in applications such as multimodal transport demand forecasting, OD matrix prediction, energy demand estimation, and environmental monitoring, where the ability to model complex spatial and temporal dependencies is critical.

The privacy-preserving constraints of FL necessitate that both the learning and testing processes are conducted independently for each client. This ensures that no sensitive information, including the scale, structure, or specifics of the data, is revealed.
These constraints are critical in maintaining the confidentiality of each client's data, which can include proprietary information and personally identifiable data.

\begin{figure}[ht]
    \centering
    \includegraphics[width=.8\linewidth]{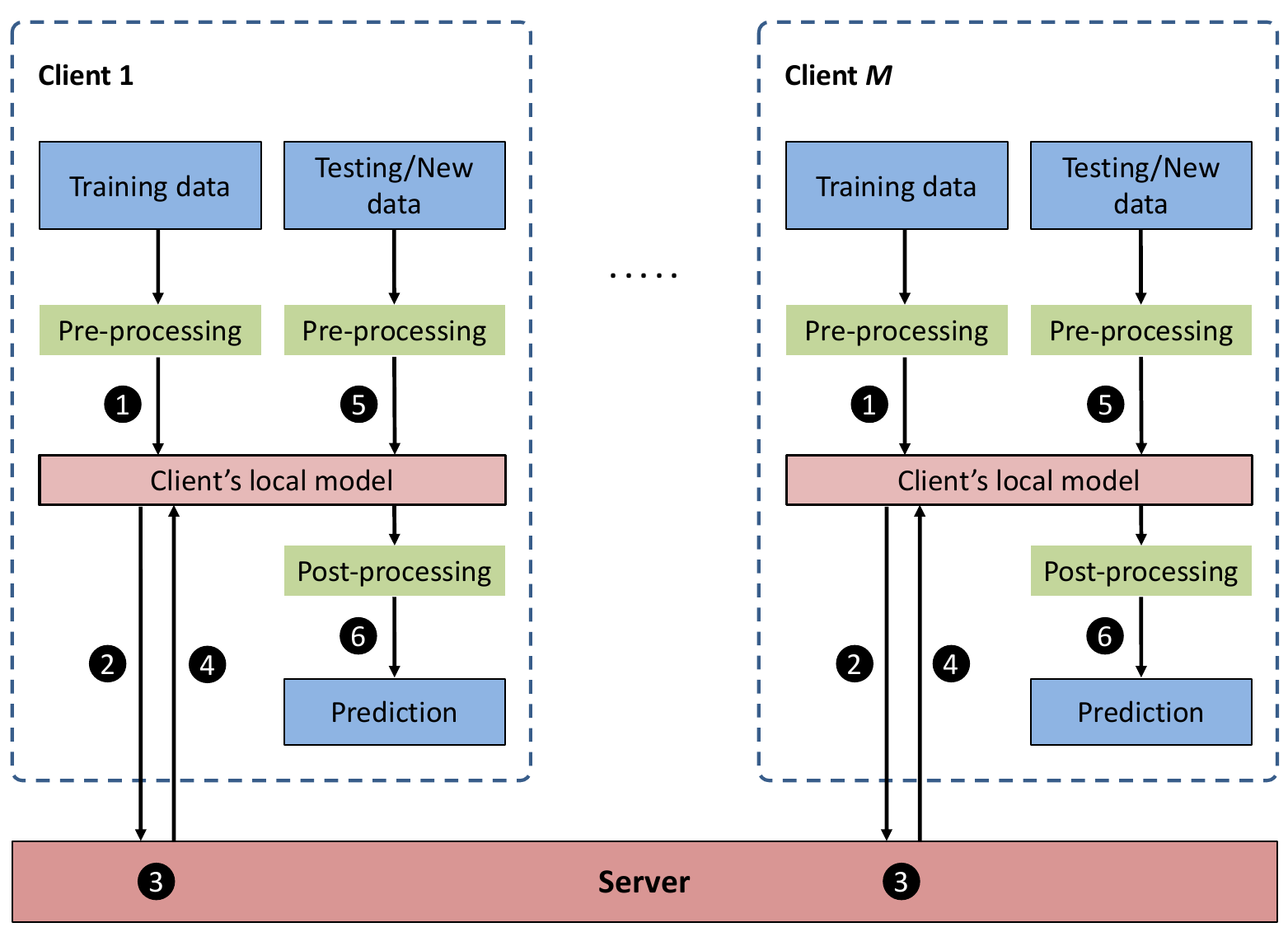}
    \caption{The workflow of FL in multimodal transport demand forecasting. {Annotation of the steps: 
    \dingg{202} client feeds pre-processed data into its local model;
    \dingg{203} client sends trained parameter to server; 
    \dingg{204} server aggregates parameters received from clients;
    \dingg{205} server sends aggregated parameter to clients;
    \dingg{206} client feeds pre-processed testing/new data into its model;
    \dingg{206} client makes prediction and transforms into the required format.}}
    \label{fig:workflow}
\end{figure}

Figure~\ref{fig:workflow} illustrates the detailed workflow of a general FL process for spatiotemporal data forecasting.
The numbered steps are as follows:
In step \dingg{202}, each client begins with their local training data, specific to their domain or region. The data is pre-processed to clean and prepare it for model training. Pre-processing steps typically include normalization, feature extraction, and handling missing values.
In step \dingg{203}, clients independently train local models using this data, optimizing parameters to fit their unique traffic patterns. After training, clients generate updates (e.g., gradients or model weights) and transmit these to a central server, ensuring privacy by not sharing raw data.
The central server aggregates these updates to form a global model (step \dingg{204}), using aggregating techniques like federated averaging (FedAvg \cite{McMahan2016FL}) or Attentive scheme (\cite{MultimodalLiLiu}) to combine local model parameters. This global model benefits from diverse data patterns, enhancing generalization.
The global model is then distributed back to all clients (step \dingg{205}) for further training with local data. 
The local training, aggregation, and global model distribution steps (steps \dingg{203} to \dingg{205}) are generally iterative and repeated over multiple rounds. During each round, clients refine their local models using the updated global model, and the process continues until the global model converges. Convergence is typically achieved when performance improvements plateau or a predefined stopping criterion is met, such as reaching a maximum number of rounds or observing no improvement in validation loss over a specified number of iterations.
The final global model is then distributed back to all clients (step \dingg{205}), with an optional final round of local training. Each client then pre-processes its new or testing data (step \dingg{206}) before using the global model to generate predictions. Lastly, the prediction outputs are post-processed and transformed into the required format (step \dingg{207}).


Although simple aggregation techniques like FedAvg have succeeded in tasks such as text prediction and image recognition, their application in transportation often yields suboptimal results. 
This is due to the significant heterogeneity among clients, not only in the number of nodes but also in the scale of data and various geographic factors. 
These differences make it challenging to create a well-performing global model, as the data characteristics in transportation are far more diverse and complex compared to more uniform tasks like text prediction or image recognition.

In this paper, we propose to enhance the FL process by introducing an additional validation step on the client side to validate the aggregated parameters received from the server before integrating them into the local model. This validation can be performed at the layer level or module level, particularly in models comprising multiple modules.
In each FL round, we aim to validate each layer/module and their combinations to find an optimal integration between the aggregated and local parameters, ensuring that the quality of the local parameters is not degraded. The details of our proposed federated scheme will be presented in Section~\ref{sec: The proposed federating schemes}.

In the next section, we present the proposed LSTM-DSTGCRN model, which serves as the local prediction models for the clients.

\section{The proposed LSTM-DSTGCRN model}
\label{sec:LSTM-DSTGCRN}

The Dynamic Spatial--Temporal Graph Convolutional Recurrent Network (DSTGCRN), introduced by \cite{GONG2024142581}, was originally developed for forecasting carbon emissions. However, this versatile model is also highly effective for multimodal transport demand prediction. 
Its strength lies in its ability to capture spatial relationships and temporal dynamics, making it well-suited for complex and evolving datasets, such as those encountered in transport forecasting. 
This model integrates three core components: a GRU, an Attention mechanism, and an AGCRN module. Each plays a pivotal role in handling the multifaceted interactions within the data.

Despite its strengths, the GRU can struggle to capture the long-term dependencies often observed in transport data. 
For example, demand patterns related to seasonal tourist inflows during summer or recurring annual festivals may exhibit trends that extend over several weeks or months, influencing transport needs long after the initial surge.
GRU's reliance on simpler gating mechanisms may limit its ability to retain and leverage information from earlier in the sequence over extended periods. 
In contrast, the LSTM has been shown to excel at capturing such long-term patterns due to its more sophisticated gating mechanisms, including the explicit memory cell that allows it to store information over longer intervals. 
To better model these temporal dependencies, we propose enhancing the architecture by replacing the GRU with an LSTM network. This enables the model to more effectively learn and predict long-term trends in transport demand data.
The resulting model will be referred to as LSTM-DSTGCRN model, and it serves as the local model for the clients in our problem. That means the local models $\mathcal F_m$ in \eqref{eq: general model} are now LSTM-DSTGCRN models, and in this section, we present the functioning of the LSTM-DSTGCRN as a forecasting model tailored for a specific client $m$. 

%

Let $\bX\in\R^{T\times N\times D}$ represent a multisource time series dataset, where $T$ is the length, $N$ is the number of nodes, and $D$ is the feature dimension. 
Let $\bX_t\in\R^{N\times D}$ denote the slice at time point $t$ of $\bX$, and $\cX_t=\{\bX_{t-p+1}, \ldots, \bX_t\}$ denote a set of $p$ recent time steps up to $t$.
In our context, as defined in Subsection \ref{subsec: Spatiotemporal data forecasting}, $\bX$ is the dataset $\sD_m$ for a specific mode of transport $m$, and $D$ consists of one dimension (usually the first) for the transport demand series and $D-1$ others for the exogenous data.
%

The LSTM-DSTGCRN model processes the multisource time series $\bX$ through a series of computational steps designed to model both the spatial and temporal dynamics inherent in the data. This proposed model unfolds into three component modules as follows.

\subsection{Long Short-Term Memory (LSTM)}

The LSTM is a type of RNN designed to address the vanishing gradient problem commonly encountered in traditional RNNs. It achieves this through a set of gating mechanisms that regulate the flow of information, allowing the network to effectively capture long-term dependencies in sequential data. 

Denote $H_t \in \mathbb{R}^{N \times d_h}$ as the hidden state and $C_t \in \mathbb{R}^{N \times d_h}$ as the cell state of the previous time step, where $d_h$ is the hidden dimension. The LSTM module in our model operates as follows:
\begin{equation}
\begin{aligned}
    F_t &= \sigma([\mathcal{X}_t, H_{t-1}] W_f + b_f), \\
    I_t &= \sigma([\mathcal{X}_t, H_{t-1}] W_i + b_i), \\
    \tilde{C}_t &= \tanh([\mathcal{X}_t, H_{t-1}] W_c + b_c), \\
    C_t &= F_t \odot C_{t-1} + I_t \odot \tilde{C}_t, \\
    O_t &= \sigma([\mathcal{X}_t, H_{t-1}] W_o + b_o), \\
    H_t &= O_t \odot \tanh(C_t),
\end{aligned}
\end{equation}
in which $F_t$ is the forget gate, $I_t$ is the input gate, $\tilde{C}_t$ is the candidate cell state, $C_t$ is the updated cell state, $O_t$ is the output gate, and $H_t$ is the new hidden state. 
The operator $\sigma$ denotes the sigmoid function, and $\odot$ denotes the Hadamard product. 
The learnable parameters include the weight matrices $W_f, W_i, W_c, W_o \in \mathbb{R}^{(D + d_h) \times d_h}$ and the bias vectors $b_f, b_i, b_c, b_o \in \mathbb{R}^{d_h}$, where $d_h$ is the hidden dimension of the LSTM network.

The flow of the LSTM module begins with the calculation of the forget gate $F_t$, which regulates the retention of information from the previous cell state $C_{t-1}$. 
This is followed by the computation of the input gate $I_t$, determining what new information will be added to the cell state. 
Concurrently, a candidate cell state $\tilde{C}_t$ is created, representing potential updates to the cell state. The updated cell state $C_t$ is then derived by combining the retained previous state and the new candidate values. 
Subsequently, the output gate $O_t$ is calculated to decide which parts of the updated cell state will contribute to the new hidden state $H_t$. 
This process allows the LSTM to effectively manage information flow, maintaining relevant data while discarding the unnecessary, thereby enhancing its capacity to model temporal dependencies in sequential data.

To introduce non-linearity, the output sequence of the LSTM network, i.e., $\cX_t' \colonequals\{H_{t-p+1}, \ldots, H_t\} \in \R^{p\times N\times d_h}$, is then passed through a linear transformation followed by a Rectified Linear Unit (ReLU) activation function as
\begin{equation}
\cX_t'' = \text{ReLU}(\cX_t' W_h  + b_h),
\end{equation}
where $W_h\in\R^{d_h\times d_e}$ and $b_h\in\R^{d_e}$ are learnable weight matrix and bias vector, respectively. 

In summary, the sequence $\cX_t''$ encodes the temporal dependencies within the input $\cX_t'$ through the LSTM module, capturing both short and long-term patterns. The LSTM's ability to manage memory through its gating mechanisms makes it particularly effective for learning sequential patterns in multimodal transport demand forecasting, or in OD matrix forecasting, where long-term dependencies, such as hourly or daily patterns, are crucial.


\subsection{Multihead attention}

Similarly as in \cite{GONG2024142581}, the multihead attention mechanism \cite{VaswaniAttention} is a pivotal component in the LSTM-DSTGCRN model, enhancing its ability to capture long-range dependencies in multimodal transport demand data. 
In the context of our model, the attention mechanism operates on the output of the LSTM module, i.e., the sequence $\cX_t''$. 
The idea is to compute the attention scores associated with each time step and spatial feature, allowing the model to focus more heavily on the most relevant elements within the data. 
This mechanism captures complex dependencies by assigning different weights to the features, highlighting the areas of the input that are more critical for predicting future transport demand.

For each attention head $k$, the mechanism computes three key vectors: the query $Q_k$, the key $K_k$, and the value $V_k$, which are derived from the input sequence $\cX_t''$ using learnable weight matrices:
\begin{align}
Q_k = \cX_t'' W^Q_k, \quad {K_k} = \cX_t'' W^K_k, \quad {V_k} = \cX_t'' W^V_k
\end{align}
where $W^Q_k$, $W^K_k$, and $W^V_k$ are learnable weight matrices. The attention scores are calculated by taking the scaled dot product of the query and key vectors, followed by applying a softmax function to normalize the weights:
\begin{align}
\text{Attention}(Q_k, K_k, V_k) = \text{softmax} \left( \frac{Q_k K_k^T}{\sqrt{d_h}} \right) V_k
\end{align}
Here, $d_h$ is the dimension of the key vectors, and the softmax function ensures that the attention scores sum to one, enabling the model to attend more to certain time steps. 


After calculating the attention scores for each head, the outputs are concatenated and passed through a linear transformation, combining the information from all attention heads:
\begin{align}
\cE_t = \text{Concat}(\text{head}_1, \dots, \text{head}_G) W^O
\end{align}
where $\text{head}_k=\text{Attention}(Q_k, K_k, V_k)$ and $W^O$ is a learnable output weight matrix, and $G$ is the number of attention heads. This final output, $\cE_t \equalscolon \{E_{t-p+1}, \ldots, E_t\} \in \R^{p\times N\times d_e}$, is then an enhanced temporal embedding representation of the input $\cX_t$.

By incorporating the multihead attention mechanism, the LSTM-DSTGCRN model effectively captures the dynamic interactions in transport data, leading to improved forecasting accuracy. 
This mechanism allows the model to adaptively prioritize different parts of the input sequence, thereby addressing the challenges of heterogeneity and complexity in the whole network of client.

\subsection{Adaptive Graph Convolutional Recurrent Network (AGCRN)}
The AGCRN model \cite{BAIYao2020} extends the capabilities of traditional graph convolutional networks (GCNs) by incorporating recurrent structures and adapting to dynamic graph structures. Similarly to  \cite{GONG2024142581}, in LSTM-DSTGCRN model, a dynamic embedding AGCRN is used to model the spatial dependencies among different locations in the transport network. The dynamic embedding AGCRN operations can be described as
\begin{equation}
\begin{aligned}
    \tilde{A} &= \text{softmax}(\text{ReLU}(E_tE_t^\top)) \\
    R_t &= \sigma( \tilde{A} [\bX_t, \tilde H_{t-1}] E_t W_r + E_t b_r ), \\
    U_t &= \sigma( \tilde{A} [\bX_t, \tilde H_{t-1}] E_t W_u + E_t b_u ), \\
    \hat H_t &= \tanh(\tilde{A} [\bX_t, U_t \odot \tilde H_{t-1}] E_t \hat W_h + E \hat b_h) \\
    \tilde H_t &= R_t \odot \tilde H_{t-1} + (1-R_t)\odot \hat H_t,
\end{aligned}
\end{equation}
where $W_r, W_u, \hat W_h \in\R^{d_e \times (D+d_h) \times d_{h}}$, and $b_r, b_u, \hat b_h \in\R^{d_e\times d_{h}}$ are learnable parameters. 
In this dynamic embedding approach, the original fixed embedding matrix in \cite{BAIYao2020} has been replaced by a dynamic one, i.e., $E_t$, which dynamically adjusts the adjacency matrix to better capture the evolving spatial dependencies. Specifically, the adjacency matrix $\tilde{A}$ is computed at each time step based on the dynamic embedding $E_t$, which encodes the temporal and spatial features of the data. This allows the model to adapt the graph structure in real time, reflecting changes in the relationships between nodes as the data evolves. For instance, during peak hours, the adjacency matrix may emphasize stronger connections between nodes with high traffic flow, while during off-peak hours, these connections may weaken. This dynamic adjustment ensures that the model remains responsive to the changing spatial dependencies inherent in transport networks.

Output of this module is a sequence $\cH_t = \{\tilde H_{t-p+1}, \ldots, \tilde H_t\} \in\R^{p\times N\times d_{h}}$ which can be viewed as an encoded version of $\cX_t$. From the last element of $\cH_t$, i.e., $\tilde H_t$, which rollingly captures all spatial and temporal information of previous time steps, we make prediction by
\begin{align}
    [\widehat Y_{i+1}, \ldots, \widehat Y_{i+Q}] = \tilde H_t \tilde W_o + \tilde b_o,
\end{align}
where $\tilde W_o\in\R^{d_{h}\times Q}$ and $\tilde b_o \in \R^Q$ are learnable weight matrix and bias vector of this final output, i.e., the prediction.

\subsection{Summary of the flow of data}

The LSTM-DSTGCRN model integrates the LSTM, Multiahead Attention mechanism, and dynamic-embedding AGCRN in a unified framework to simultaneously capture temporal and spatial dependencies in multimodal transport data. The workflow of the LSTM-DSTGCRN can be summarized as follows:

\begin{enumerate}[label=\it \roman*.]
    \item Input processing: Transport demand data is preprocessed and fed into the LSTM to capture temporal dependencies.
    \item Attention mechanism: The output of the LSTM is passed through the multiahead attention mechanism to dynamically weight the importance of different time steps and locations.
    \item Spatial modeling: The weighted outputs are then processed by the AGCRN with dynamic embedding to capture spatial dependencies and interactions across the transport network.
    \item Prediction: The final output of AGCRN is used to predict future transport demand via a linear transformation.
\end{enumerate}
The list of the LSTM-DSTGCRN model's parameters is provided in Table~\ref{tab:parameters} for reference.
\begin{table}[H]
    \centering
    \caption{LSTM-DSTGCRN model parameters}
    \label{tab:parameters}
    \begin{tabular}{lll}
        \toprule
        Module & Parameters & Dimension \\
        \midrule
        LSTM & $W_f, W_i, W_c, W_o$ & $(D + d_h) \times d_h$ \\
        & $b_f, b_i, b_c, b_o$ & $d_h$ \\
        Multihead Attention & $W^Q_k, W^K_k, W^V_k$ & $d_h \times d_h$ \\
        & $W^O$ & $Gd_h \times d_e$ \\
        AGCRN & $W_r, W_u, \hat W_h$ & $d_e \times (D + d_h) \times d_{h}$ \\
        & $b_r, b_u, \hat b_h$ & $d_e\times d_{h}$ \\
        Final Output & $\tilde W_o$ & $d_{h} \times Q$ \\
        & $\tilde b_o$ & $Q$ \\
        \bottomrule
    \end{tabular}
\end{table}

In the subsequent sections we can see that integration into a FL framework further enhances our model applicability by enabling collaborative learning across distributed datasets while preserving data privacy.

\section{The proposed federating schemes}\label{sec: The proposed federating schemes}


\subsection{Client-Side Validation (CSV) mechanism}
As introduced in Subsection \ref{subsec: Federated learning in spatiotemporal data forecasting}, our proposed FL scheme incorporates a CSV mechanism to enhance model robustness and accuracy. 
This novel scheme introduces an additional validation step on the client side before updating the local model with the aggregated parameters received from the server, ensuring that only the most beneficial updates are integrated into the local model.

Let $\mathcal{Z}$ denote the set of indices of layers (or modules, depending on the desired level of validation detail) in the architecture of the model $\mathcal{F}$. The power set of $\mathcal{Z}$ is denoted by $\mathcal{P}(\mathcal{Z})$. For each $S \in \mathcal{P}(\mathcal{Z})$, where $S$ represents a subset of indices of layers or modules, let $\bsP_{m,S}$ represent the portion of $\bsP_m$ corresponding to $S$.
In our proposed FL scheme, at each FL round, each client has the possibility of a partial update of its layer/module weights.
The proposed FL approach is detailed in Algorithm \ref{algo: main Fed-DSTGCRN}.

\begin{algorithm}[H]
\caption{Pseudo-code of Fed-LSTM-DSTGCRN model.}\label{algo: main Fed-DSTGCRN}
\begin{algorithmic}[1]
\vspace{0.1cm}
\STATEx{\textbf{Input:} $\sD_m$: dataset of client $m$ ($m=1,\ldots,M$).}
\STATEx{\textcolor{white}{\textbf{Input:}} $\mathcal{F}$: architecture for LSTM-DSTGCRN model (shared by all clients).}
\STATEx{\textcolor{white}{\textbf{Input:}} $R_{\text{max}}$: maximum number of FL rounds.}
\STATEx{\textcolor{white}{\textbf{Input:}} $E_m$: number of epochs for local training.}
\STATEx{\textbf{Output:} $\bsP^*_m$: best performing parameter for each client $m$ ($m=1,\ldots,M$).}
\vspace{0.15cm}\hrule\vspace{0.15cm}
\STATE{Initialize $\widetilde\bsP^{(0)}$ for $\mathcal{F}$.}
\STATE{Distribute $\mathcal{F}$ and $\widetilde\bsP^{(0)}$ to all clients.}
\STATE{$r \longleftarrow 0$ (index for FL round).}
\STATE{Initialize $L^*_m \gets \infty$, $\forall m = 1, \ldots, M$ (large initial value to ensure proper update).}
\REPEAT
    \FOR{$m=1,\ldots,M$ (possibly in parallel)}
        \STATE{\textit{(Receiving)} Client $m$ receives $\widetilde \bsP^{(r)}$ from server.}
        \STATE{\textbf{if} $r=0$ \textbf{then} $\bsP^{(0)}_m \longleftarrow \widetilde\bsP^{(0)}$ \textbf{end if}}\COMMENT{Initializing, only for the first round.}
        \STATE{\textit{(Validating)} Client $m$ validates $\widetilde\bsP^{(r)}$ using Algorithm \ref{Client-Side Validation algorithm}.} 
        \STATE{$\widetilde\bsP^{(r)}_{m} \longleftarrow$ integrated parameter after validation.}
        \STATE{\textit{(Training)} Client $m$ trains $\mathcal{F}$ with $\widetilde\bsP^{(r)}_{m}$ as the starting point for $E_m$ epochs.}
        \STATE{$\bsP^{(r)}_{m} \longleftarrow$ obtained parameter after training.}
        \STATE{\textit{(Sending)} Client $m$ sends $\bsP^{(r)}_{m}$ to server for aggregation.}
    \ENDFOR
    \STATE{\textit{(Collecting)} Server collects $\bsP^{(r)}_{m}$ from all clients $m$.}
    \STATE{\textit{(Aggregating)} $\widetilde\bsP^{(r+1)} \longleftarrow \text{Aggregate}(\bsP^{(r)}_{1}, \ldots, \bsP^{(r)}_{M})$ using FedAvg scheme.}
    \STATE{\textit{(Distributing)} Server distributes $\widetilde\bsP^{(r+1)}$ to all clients.}
    \STATE{$r \longleftarrow r+1$.}
\UNTIL{$r \geqslant R_{\text{max}}$ or early stopping condition met.}
\STATE{\textbf{return} $\bsP^*_m$ for each client $m$, $m=1,\ldots,M$.}
\end{algorithmic}
\end{algorithm}

Specifically, the process begins with the initialization of the model architecture $\mathcal{F}$ and an initial weight $\widetilde\bsP^{(0)}$ on the server side. This initial weight, along with the model architecture, are distributed to all participating clients.

The FL process proceeds iteratively over a series of rounds, indexed by $r$. During each round, clients receive the current global parameters $\widetilde\bsP^{(r)}$ from the server. In the first round ($r=0$), clients initialize their local parameters $\bsP^{(0)}_m$ to match the received global parameters.

A crucial step of this FL approach is the CSV mechanism (line 9 in Algorithm \ref{algo: main Fed-DSTGCRN}). Upon receiving the global parameters, each client temporarily integrates them into its local model and validates these parameters using a local validation dataset $\sD^{\val}_m$. 
More specifically, for each $S \in \mathcal{P}(\mathcal{Z})$, client computes the corresponding validation loss $\mathcal{L}(\bsP^{(r)}_{m,S}; \sD^{\val}_m)$,
then find the optimal integration $\bsP^{(r)}_{m,S^*}$ to continue the training. This procedure is described in Algorithm \ref{Client-Side Validation algorithm}.
\begin{algorithm}[!h]
\caption{Client-Side Validation algorithm}\label{Client-Side Validation algorithm}
\begin{algorithmic}[1]
\vspace{0.1cm}
\STATEx{\textbf{Input:} $\sD^{\val}_m$: validation data of client $m$.}
\STATEx{\textcolor{white}{\textbf{Input:}} $\bsP^{(r)}_{m}$: local parameter after local training at round $r$.}
\STATEx{\textcolor{white}{\textbf{Input:}} $\widetilde\bsP^{(r)}$: aggregated parameter received from the server.}
\STATEx{\textbf{Output:} $\widetilde\bsP_{m}^{(r)}$: optimal integration for client $m$ in subsequent learning.}
\vspace{0.15cm}\hrule\vspace{0.15cm}
\STATE{Let $\mathcal{Z}$ denote the set of layers (or modules) of $\bsP^{(r)}_{m}$.}
\FOR{\textbf{each} $S \in \mathcal{P}(\mathcal{Z})$}
    \STATE{Replace $S$ in $\bsP^{(r)}_{m}$ with corresponding part in $\widetilde\bsP^{(r)}$, yielding $\bsP^{(r)}_{m,S}$ (a candidate for integration).}
    \STATE{Compute $\mathcal L(\bsP^{(r)}_{m,S}; \sD^{\val}_m)$, validation loss for client $m$ in case using $\bsP^{(r)}_{m,S}$.}
\ENDFOR
\STATE{Find the optimal integration $\bsP^{(r)}_{m,S^*} = \arg\min_{S \in \mathcal{P}(\mathcal Z)} \mathcal L(\bsP^{(r)}_{m,S}; \sD^{\val}_m)$.}
\STATE{$\widetilde\bsP^{(r)}_{m} \longleftarrow \bsP^{(r)}_{m,S^*}$.}
\STATE{\textbf{return} $\widetilde\bsP^{(r)}_{m}$}
\end{algorithmic}
\end{algorithm}

\begin{figure}[!h]
    \centering
    \includegraphics[width=.8\linewidth]{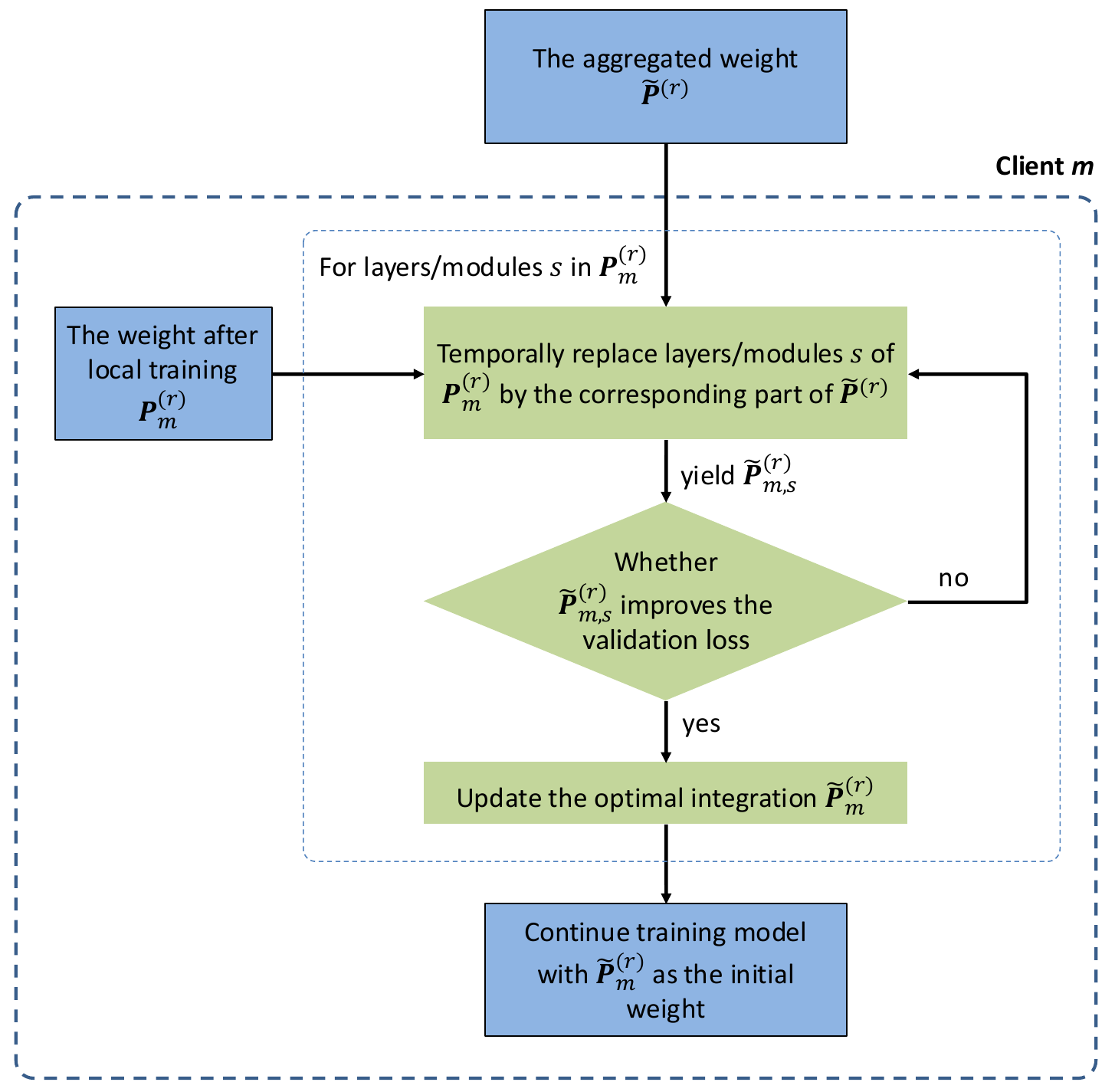}
    \caption{Flow of the Client-Side Validation  mechanism.}
    \label{fig:CSV_diagram}
\end{figure}

A diagram of the proposed CSV mechanism is shown in Figure~\ref{fig:CSV_diagram}.

Following validation, the client uses the integrated parameters $\widetilde\bsP^{(r)}_{m}$ as the starting point for local training over a specified number of epochs $E_m$. The updated parameters $\bsP^{(r)}_{m}$, resulting from this local training, are then transmitted back to the server.

The server collects the updated parameters from all clients and aggregates them using Federated Averaging (FedAvg) scheme to generate the new global parameters $\widetilde\bsP^{(r+1)}$. These updated parameters are then distributed to all clients, and the process repeats for the next round.

This iterative FL process continues until the maximum number of rounds $R_{\text{max}}$ is reached, or an early stopping condition is met. 
Upon completion, the algorithm returns the best-performing parameters $\bsP^*_m$ for each client, ensuring that each client benefits from the FL process while maintaining the privacy of their local data. The CSV step plays a pivotal role in optimizing model performance across diverse and heterogeneous data sources.


\subsection{Proposed FL scheme tailored for LSTM-DSTGCR model}
AS discussed, the LSTM-DSTGCR model has three main modules: LSTM, Multihead Attention, and AGCRN.
Applying the proposed CSV for the LSTM-DSTGCRN model at module level results in an approach that will be referred to as ``LSTM-DSTGCRN + FedAvg + CSV''.
In summary, this approach consists of the following steps:
\begin{enumerate}
    \item \textbf{Initialization}: The server initializes the global LSTM-DSTGCRN model and distributes it to all clients.
    \item \textbf{Local training}: Each client trains the local LSTM-DSTGCRN model on its dataset for some local epochs.
    \item \textbf{Parameter sharing}: Clients send their locally trained model parameters to the server.
    \item \textbf{Server-side aggregation}: The server aggregates the received parameters using FedAvg scheme.
    \item \textbf{Client-side validation}: Upon receiving the aggregated parameters, each client performs the following steps for each subset of the three modules of the LSTM-DSTGCRN:
    \begin{enumerate}
        \item \textbf{Temporary update}: Replace the local modules parameters with the corresponding aggregated parameters.
        \item \textbf{Validation}: Compute the validation loss using a local validation set.
        \item \textbf{Selective update}: If the validation loss improves, retain the aggregated parameters; otherwise, revert to the original local parameters.
    \end{enumerate}
    \item \textbf{Global update}: The validated and selectively updated local models are used for the next round of local training.
\end{enumerate}

By validating the updates, the framework leverages the strengths of each component while mitigating the impact of any suboptimal parameter updates.

\section{Experiments}
\label{sec:experiments}

In this section, we present the experiments conducted to evaluate the performance of the LSTM-DSTGCRN model within the proposed FL framework. We begin by describing the datasets used in the study, followed by the experimental setup and the evaluation metrics employed.
A comprehensive analysis of the results, including a comparison of the LSTM-DSTGCRN model with other state-of-the-art models (with and without the proposed FL scheme), an ablation study, and the impacts of number of FL rounds, is presented in the subsequent sections.

\subsection{Datasets}

\subsubsection{Transport demand datasets}
For the application of multimodal transport demand forecasting, we utilize three real-world public datasets, which include bike and taxi demand data from New York City (NYC) and Chicago (CHI), as described in Table \ref{tab: datasets}.

\begin{table}[H]
    \centering
    \caption{Datasets description}
    \label{tab: datasets}
    \begin{tabular}{lccc}
        \toprule
        Dataset & Period & Number of nodes  \\
        \midrule
        NYC-Bike\tablefootnote{\url{https://citibikenyc.com/system-data}} 
        & From 01/04/2016 to 30/06/2016 & 283 \\
        NYC-Taxi\tablefootnote{\url{https://www.nyc.gov/site/tlc/about/tlc-trip-record-data.page}} 
        & From 01/04/2016 to 30/06/2016 & 263 \\
        CHI-Taxi\tablefootnote{\url{https://data.cityofchicago.org/Transportation/Taxi-Trips-2024/sa9s-wkhk/}}  
        & From 01/04/2024 to 30/06/2024 & 77 \\
        \bottomrule
    \end{tabular}
\end{table}

The datasets were aggregated \st{into an hourly timeframe} {at an hourly resolution}, resulting in 2184 observations for each node. 
Our hypothesis is that the transport demand patterns from the datasets can mutually enhance each other, whether they represent the same mode of transportation (e.g., NYC-Taxi and CHI-Taxi) or operate within the same city (e.g., NYC-Taxi and NYC-Bike). 
In our experiments, we forecast the demand for the next hour based on the 12 most recent demand observations.

\subsubsection{OD matrix datasets}


For the application of OD matrix forecasting, we utilize two private datasets capturing trip patterns in the Lyon urban area, the second-largest city in France by population. The data was collected from two distinct entities: the \textit{Criter System}, an automated road traffic regulation system managed by Greater Lyon's Highways Department, and \textit{Orange Telecom}, a leading telecom operator in France. 
Our objective is to investigate whether collaboration between these two entities can enhance the forecasting of OD trips for each. For simplicity, we refer to these entities as Lyon PT (stands for public transport) and Orange Telecom, respectively.

The data consists of OD matrices aggregated at 2-hour intervals, except during nighttime (1 AM to 7 AM), where a single aggregation is used. As a result, each day is divided into 10 time slots, and the OD matrix of each slot represents the number of passengers traveling between different locations within the city during that period. 
The OD data covers six zones: \textit{96091}, \textit{96092}, \textit{96093}, \textit{96094}, \textit{Lyon} (an aggregation of the five zones that make up the city of Lyon), and \textit{Outside Urban Area} (an aggregation of 77 zones located outside the Lyon urban area). The number of OD pairs (i.e., nodes) is therefore $6 \times 6 -6  = 30$ for each time slot, in which the ``$-6$'' is due to the exclusion of self-loops.

The data was collected for March and September in two separate years: 2021 and 2022. Due to the impact of the Covid-19 pandemic in France, mobility patterns in 2021 differ significantly from those in 2022, prompting us to analyze the two years separately.

\subsection{Experiment setup}

\textit{Data pre-processing}: The pre-processing of datasets on local machines is a crucial step to ensure that the data is appropriately prepared for the FL framework. On each local machine, the dataset is first split into three subsets: a training set (70\%), a validation set (20\%), and a testing set (10\%). Since OD data was collected in two non-consecutive months, we apply the splitting to each month separately (for each year).

Following the data split, the training and validation sets are standardized to follow a normal distribution, which is essential for improving the convergence of the model during training. 
To maintain the fairness and realism of a real-world scenario, the mean and covariance computed from the training set are stored and later used to standardize the testing set, as well as to de-standardize the predictions to obtain the actual values.
This approach ensures that the testing set is standardized in a manner consistent with the data that the model was trained on, preventing data leakage and ensuring the integrity of the evaluation process.

To more effectively capture the underlying temporal patterns in the data, we incorporated additional features by encoding the timeslot of the day and the day of the week, which allows the model to recognize and exploit daily and weekly patterns. 
Additionally, for transport demand datasets, hourly temperature ($^{\circ}\text{C}$) and precipitation (mm) were also utilized for predictions, as weather is known to significantly impact transportation. Specifically, for each dataset, we collected hourly weather data in the corresponding periods in Table \ref{tab: datasets} from Open-Meteo\footnote{\url{https://open-meteo.com/}} platform for all nodes based on their coordinates. 
So, in addition to the primary transport demand time series, each node includes four additional exogenous time series, resulting in a total feature dimension of five for the final inputs.
By incorporating these external factors, we provide the model with richer input, enabling it to better understand and forecast the complex behaviors inherent in transport demand data.

\textit{Technical settings}: To accurately replicate real-world scenarios, we simulate each local machine as an independent Python process, with a dedicated server process orchestrating the FL process. 
This setup enables us to closely mimic the decentralized nature of FL, where each client operates independently while coordinating with the central server.

All experiments were conducted on a computing setup equipped with a NVIDIA RTX 3090 GPU, with 24 GB of memory, and a system RAM of 32 GB.
The experiments were run using Python 3.11, with CUDA 12.2 and PyTorch 2.4.0 handling the computational backend.
 {We select the initial learning rate based on model complexity. For standard time-series models without graph components (i.e., GRU, LSTM and their federated counterparts), we use an initial learning rate of $0.001$. 
For graph-based models and variants that include graph components (i.e., AGCRN, DSTGCRN, LSTM-DSTGCRN and their federated counterparts), we use a smaller initial learning rate of $0.0001$ to improve training stability.
For the main results, we report runs with batch size $16$ across methods for a fair comparison.
For FL experiments in the main text, we used $50$ FL rounds with $6$ local epochs per round for both transport demand datasets and OD-matrix datasets.
For centralized training (i.e., local learning without federation), we implemented early stopping with a patience of 10 epochs based on validation loss. For FL approaches, we monitored validation performance after each FL round as well as after each local epoch within rounds, and we restored the model weights corresponding to the best validation performance after each FL round for final evaluation on the test set.}
 {Finally}, to ensure transparency and reproducibility, we seeded the random number generator with a conventional value of 42.

\subsection{Metrics}\label{subsec:Metrics}
In our experiments, we evaluate the performance of the proposed model using two widely recognized metrics: Mean Absolute Error (MAE) and Root Mean Squared Error (RMSE). These metrics are defined as follows:
\begin{align*}
    \text{MAE} &= \frac{1}{n} \sum_{i=1}^{n} \left| y_i - \hat{y}_i \right|,\\
    \text{RMSE} &= \sqrt{\frac{1}{n} \sum_{i=1}^{n} \left( y_i - \hat{y}_i \right)^2},
\end{align*}
where $y_i$ is the actual value, $\hat{y}_i$ is the predicted value, and $n$ is the number of predictions made. 
MAE measures the average magnitude of errors in a set of predictions, without considering their direction, provides a straightforward interpretation of the average prediction error. 
Whereas, RMSE gives more weight to larger errors, making it sensitive to outliers, it is particularly useful when the consequences of large errors are significant.

\section{Results and Discussion}
\label{sec:results discussion}

In this section, we present the results of our experimental studies conducted on the real-world datasets described earlier (the involving application will be mentioned in the titles of the subsections). We provide a comparison between the local and FL models and discuss the performances of the LSTM-DSTGCRN model with the proposed FL scheme. 
We also provide some insights into the training process and present the results of the ablation study, which aims to analyze the contributions of individual components within the LSTM-DSTGCRN model. Finally, we investigate the impact of the number of FL rounds on the model's performance and discuss the limitations of the proposed FL scheme.

\subsection{Comparison of local models and FL approaches on transport demand datasets}\label{subsec: local models transport demand datasets}

In this study, we first benchmark our proposed LSTM-DSTGCRN model against several baseline models commonly used in time series analysis and transport demand forecasting. The baselines include:

\begin{itemize}
\item Gated Recurrent Unit (GRU): In this baseline, each dataset is treated as a set of multivariate time series, with the GRU model employed to learn temporal dependencies and forecast future values.
\item Long Short-Term Memory (LSTM): Similar to GRU, the LSTM model is applied to each multivariate time series dataset to capture long-term dependencies, serving as another important deep learning baseline.
\item Adaptive Graph Convolutional Recurrent Network (AGCRN) \cite{BAIYao2020}: An advanced model that combines graph convolutional networks with recurrent neural networks, enabling the capture of complex spatiotemporal dependencies in graph-structured data, making it particularly effective for transport demand forecasting.
\item Dynamic Spatial--Temporal Graph Convolutional Recurrent Network (DSTGCRN) \cite{GONG2024142581}: The original model upon which our local model is based, combining graph convolutional and recurrent layers to model dynamic spatiotemporal data.
\item LSTM-DSTGCRN: Our proposed model for local training enhances the DSTGCRN by replacing the GRU with LSTM networks. This modification improves the model's ability to capture long-term dependencies, making it better suited to the complex requirements in multimodal transport demand forecasting.
\end{itemize}
By comparing the performance of our model against these baselines, we can rigorously evaluate its effectiveness in handling the complexities of transport demand forecasting across diverse scenarios.

Secondly, we conduct a comparison of various FL approaches to evaluate their effectiveness in multimodal transport demand forecasting. 
The approaches compared include:
\begin{itemize} 
\item FedGRU \cite{9082655}: A FL variant of the GRU model, where each client independently trains a GRU model on local data. The locally trained weights are then aggregated on the server using the FedAvg scheme, which helps to enhance overall model performance while maintaining data privacy across clients. 
\item FedLSTM \cite{Zeng2021MultiTaskFL}: A federated adaptation of the LSTM network, designed to capture long-term dependencies in sequential data. Similar to FedGRU, the FedLSTM approach aggregates locally trained LSTM models at the server, allowing for improved forecasting performance without compromising the privacy of individual clients' data. 
\item AGCRN + Attentive \cite{MultimodalLiLiu}: A variant of the AGCRN model \citep{BAIYao2020} that incorporates the Attentive FL scheme proposed by \cite{MultimodalLiLiu}, where each client is weighted based on its similarity to others. This allows for a more tailored aggregation process, improving the model's ability to generalize across different client datasets. 
\item LSTM-DSTGCRN + Attentive: The proposed LSTM-DSTGCRN model with the Attentive FL scheme proposed by \cite{MultimodalLiLiu}.
\item LSTM-DSTGCRN + FedAvg: The proposed LSTM-DSTGCRN model with FedAvg scheme for all modules.
\item LSTM-DSTGCRN + FedAvg with CSV: Our proposed FL scheme that integrates a CSV mechanism at the module level, ensuring that only the most beneficial updates are applied. This approach balances collaboration with individual client needs, enhancing the robustness and accuracy of the overall model. 
\end{itemize}

\begin{table*}[!h]
    \centering
{\footnotesize
    \caption{Performance comparison of local models on transport demand datasets}
    \label{tab:comparison of local models}
    \begin{tabular}{lcccccc}
        \toprule
        \multirow{2}{*}{Model} & \multicolumn{2}{c}{NYC-Bike} & \multicolumn{2}{c}{NYC-Taxi} & \multicolumn{2}{c}{CHI-Taxi} \\
        \cmidrule(r){2-3} \cmidrule(r){4-5} \cmidrule(r){6-7}
        & MAE & RMSE & MAE & RMSE & MAE & RMSE \\
        \midrule
        GRU & 2.1518 & 3.3612 & 11.7579 & 31.6111 & 2.6852 & 6.9351 \\
        LSTM & 1.9727 & 3.2877 & 11.3981 & 30.5406 & 2.9644 & 7.0128 \\
        AGCRN \cite{BAIYao2020} & 1.9793 & \textbf{2.9422} & 10.3327 & \textbf{24.1955} & 2.8162 & 6.8642 \\
        DSTGCRN \cite{GONG2024142581} & 1.9593 & 3.0953 & 9.9527 & 26.4385 & 2.4335 & 6.6989 \\
        LSTM-DSTGCRN (ours) & 1.8994 & 3.0450 & 9.8571 & 26.3968 & 2.6110 & 6.6846 \\
        \midrule
        FedGRU \cite{9082655} & 2.4897 & 3.5930 & 11.0238 & 31.4011 & 2.7614 & \textbf{6.6305} \\
        FedLSTM \cite{Zeng2021MultiTaskFL} & 2.2396 & 3.4365 & 11.5929 & 31.2145 & 2.9620 & 7.4549 \\
        AGCRN + Attentive \cite{MultimodalLiLiu} & 2.0143 & 3.1892 & 11.0276 & 27.3424 & 3.1452 & 8.2231 \\
        LSTM-DSTGCRN + Attentive & 1.9323 & 3.2024 & 11.2389 & 26.1926 & 2.5551 & 6.8292 \\
        LSTM-DSTGCRN + Attentive + CSV & \textbf{1.8667} & 3.4377 & 11.7614 & 28.3069 & \textbf{2.3127} & 8.2746 \\
        LSTM-DSTGCRN + FedAvg  & 1.9423 & 3.3253 & 10.3052 & 26.8824 & 3.3121 & 6.9915 \\
        LSTM-DSTGCRN + FedAvg with CSV  & 1.8677 & 3.1978 & \textbf{9.3249} & 26.1514 & 2.5453 & 7.2776 \\
        \bottomrule
    \end{tabular}
}\end{table*}
{Note that, the model DSTGCRN \citep{GONG2024142581} can be viewed as a GRU-DSTGCRN variant. While other recurrent architectures such as TCN could potentially enhance model performance, we restrict our analysis to GRU and LSTM-based implementations to maintain experimental focus.}

The overall comparison can be seen in Table \ref{tab:comparison of local models}. This table is read as follows: above the middle line are the models that are trained locally by each client (using their data only), while below are the FL approaches. The smallest MAE and RMSE values are highlighted in bold. The lower the MAE and RMSE values, the better the model's performance.

Considering the local models, i.e., above the middle line, the results indicate that our proposed LSTM-DSTGCRN performs competitively, often matching or slightly surpassing the baselines. 
On the NYC-Bike and NYC-Taxi datasets, LSTM-DSTGCRN achieves the best MAEs (1.8994 and 9.8571, respectively), whereas AGCRN has the lowest RMSE (2.9422 and 24.1955, respectively). This suggests that, in New York city, while the LSTM-DSTGCRN excels in capturing long-term dependencies, the AGCRN is more effective in reducing prediction variance.
For the CHI-Taxi dataset, the DSTGCRN has the lowest MAE (2.4335), but the LSTM-DSTGCRN closely follows with an MAE of 2.6110 and the best RMSE (6.6846), demonstrating its robustness in reducing prediction variance. These results suggest that the LSTM-DSTGCRN provides a balanced approach to capturing both long-term temporal and spatial dependencies, showing modest improvements over the original DSTGCRN and other baselines across different datasets.

{We observe that the MAE–RMSE divergence is pronounced for LSTM-DSTGCRN + Attentive (+CSV) but much less evident for LSTM-DSTGCRN + FedAvg (+CSV). This difference can be explained from the aggregation mechanism point of view. Attentive aggregation reweights each client's layer updates by how similar they are to other clients. 
Updates coming from clients that look more alike receive higher weights, while more distinctive clients are downweighted. Combined with CSV, which only accepts modules that improve validation loss, this drives the model toward the high-density, ``typical'' behavior shared by many clients, results in shrinking the MAE. 
Whereas, for rare, event-driven demand spikes where some clients' patterns genuinely diverge, the similarity weighting can dilute those locally critical signals. The result can be occasional larger residuals on tail cases, which shows up as a higher RMSE despite an improved MAE.
By contrast, FedAvg averages client updates more uniformly and thus does not suppress atypical clients as strongly. When combined with CSV, it continues to reduce frequent errors while preserving more signal from rare regimes, which results in the MAE–RMSE gap occurring less frequently (only on CHI-Taxi) compared to Attentive+CSV.}

\begin{figure}[!h]
    \centering
    \includegraphics[width=0.7\linewidth]{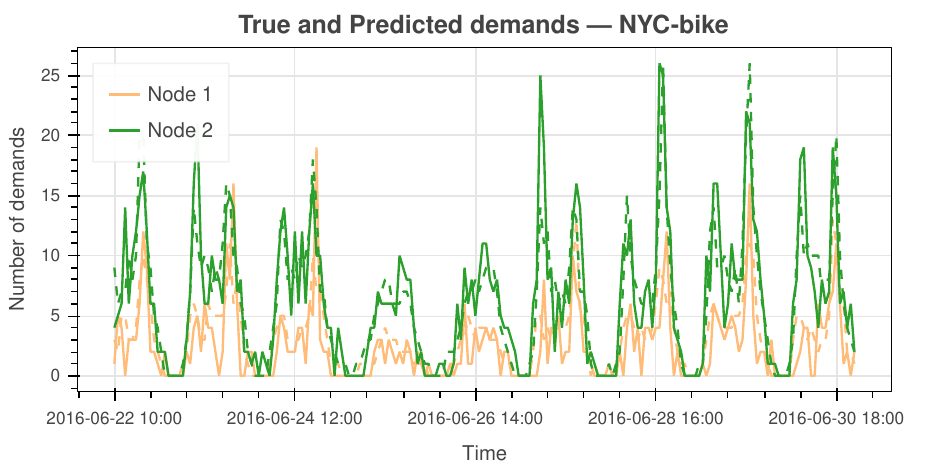}\\
    \includegraphics[width=0.7\linewidth]{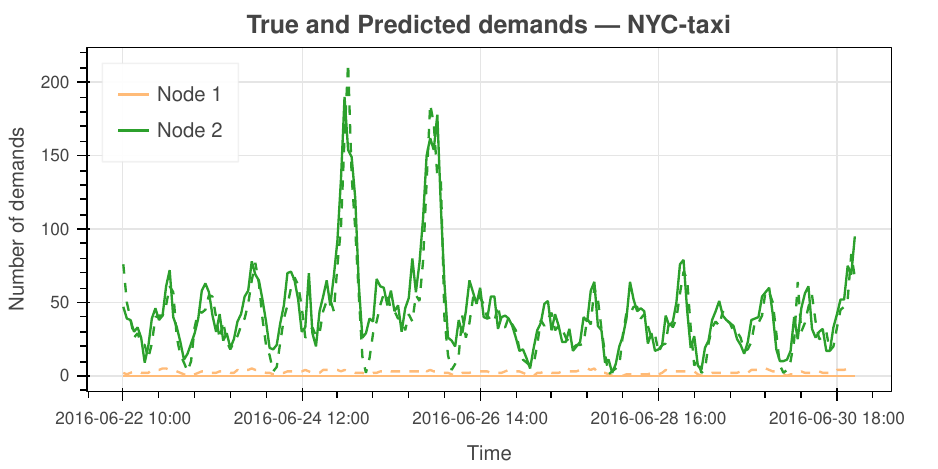}\\
    \includegraphics[width=0.7\linewidth]{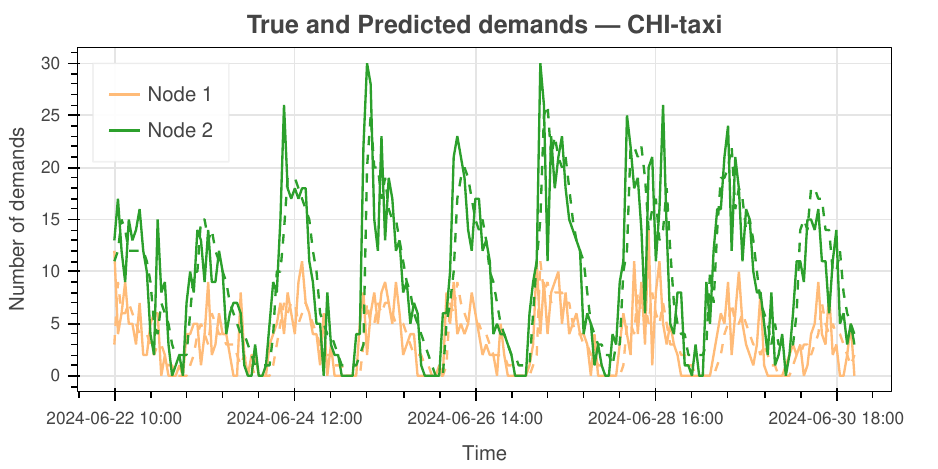}
    \caption{Forecasts given by LSTM-DSTGCRN at some randomly chosen nodes of the transport demand datasets. The dashed lines present the predicted values, the solid lines present the true values.}
    \label{fig:predictions}
\end{figure}

Figure \ref{fig:predictions} visualize the forecasts given by LSTM-DSTGCRN at some random nodes for each of the transport demand dataset.

Considering the FL approaches, i.e., below the middle line, the results demonstrate that the proposed CSV mechanism, either with Attentive or FedAvg schemes, significantly improves the MAEs on all datasets. For example, on NYC-Bike and CHI-Taxi datasets, the using of CSV mechanism helped the LSTM-DSTGCRN + Attentive and LSTM-DSTGCRN + FedAvg approaches to achieve better MAEs. These results highlight the effectiveness of the proposed CSV mechanism in enhancing the robustness and accuracy of the LSTM-DSTGCRN model.

{Overall}, on NYC-Bike, LSTM-DSTGCRN + Attentive + CSV gives the best MAE (1.8667), while LSTM-DSTGCRN + FedAvg + CSV follows closely with an MAE of 1.8677, both of them beat all of the local trained models, i.e., without FL. 
On NYC-Taxi, LSTM-DSTGCRN + FedAvg + CSV achieves the best MAE (9.3249) and RMSE (26.1514), outperforming all other FL approaches and local models. On CHI-Taxi, LSTM-DSTGCRN + FedAvg + CSV also achieves the best MAE (2.3127), and on CHI-Taxi, LSTM-DSTGCRN + Attentive + CSV achieves the best MAE (2.3127), outperforming all other FL approaches and local models. 
These results highlight the effectiveness of the FL combined with CSV mechanism. In other words, with FL, the clients can collaboratively enhance the model's performance, without compromising data privacy, while the usage of CSV mechanism ensures that only beneficial updates are integrated, thereby improving the overall model performance.

\subsection{Insights from the training process on transport demand datasets}
\begin{figure}[H]
    \centering
    \includegraphics[width=0.31\linewidth]{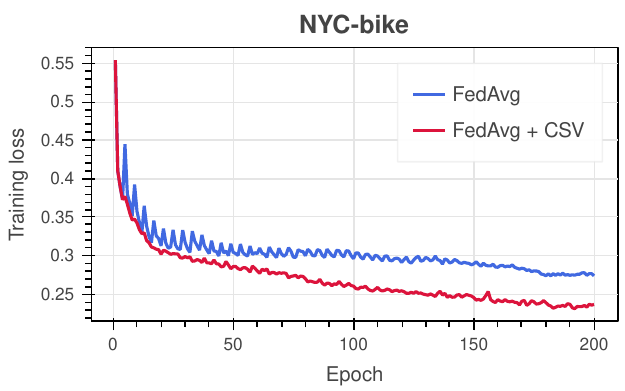}
    \includegraphics[width=0.31\linewidth]{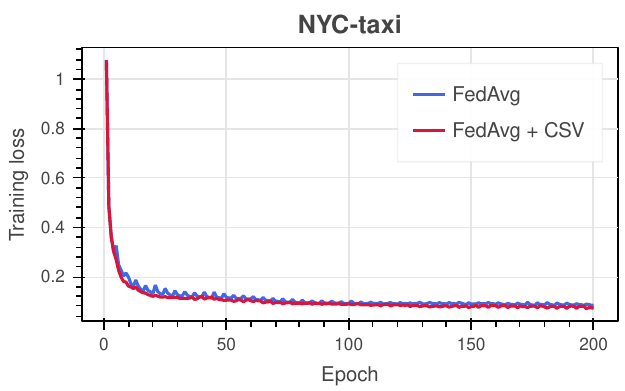}
    \includegraphics[width=0.31\linewidth]{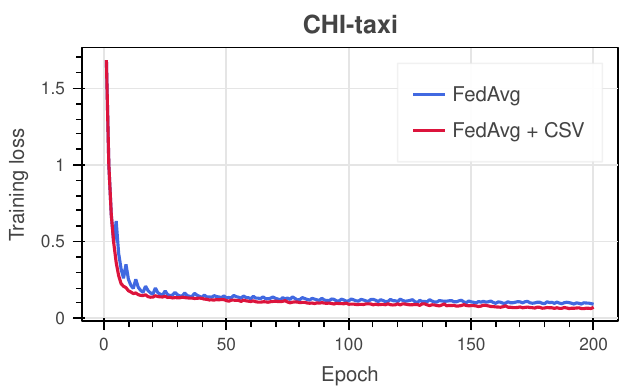}
    \caption{Training losses of FL for LSTM-DSTGCRN with FedAvg and FedAvg+CSV.}\label{fig:fig-loss}
\end{figure}

Figure \ref{fig:fig-loss} shows the training losses of the two approaches LSTM-DSTGCRN + FedAvg and LSTM-DSTGCRN + FedAvg + CSV. We can see that the direct integration of aggregated parameters without CSV results in greater fluctuations in training losses. 
In contrast, the CSV mechanism stabilizes the training process, leading to more consistent and lower losses. This suggests that the CSV mechanism effectively filters out noisy updates, ensuring that only beneficial information is integrated into the local models. This results in more stable and accurate predictions, as evidenced by the improved performance of the LSTM-DSTGCRN + FedAvg + CSV approach over LSTM-DSTGCRN + FedAvg.

\begin{figure}[!h]
    \centering
    \includegraphics[width=0.6\linewidth]{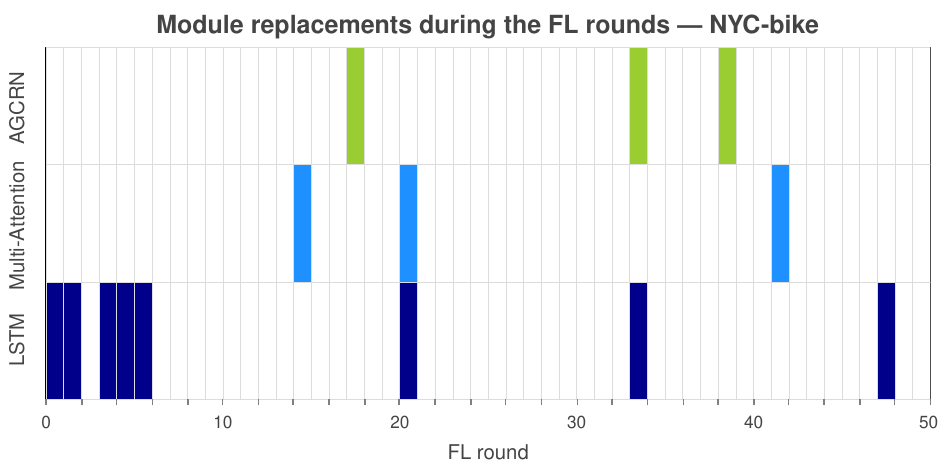}\\
    \includegraphics[width=0.6\linewidth]{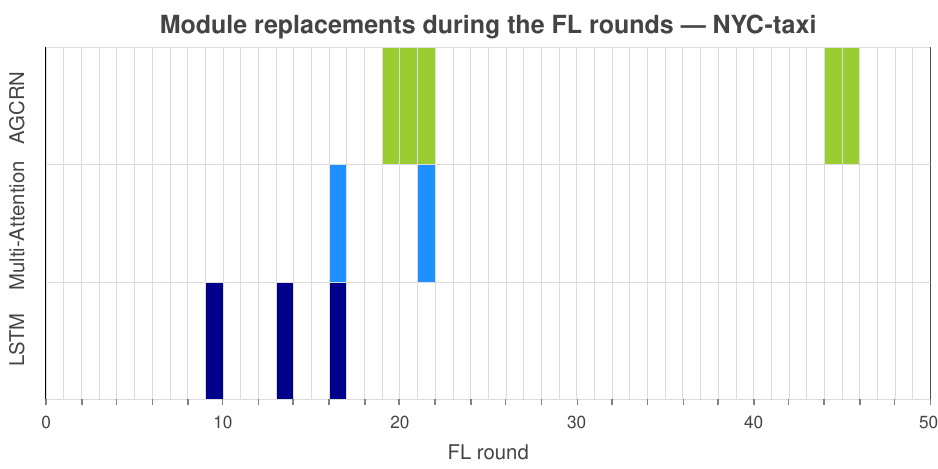}\\
    \includegraphics[width=0.6\linewidth]{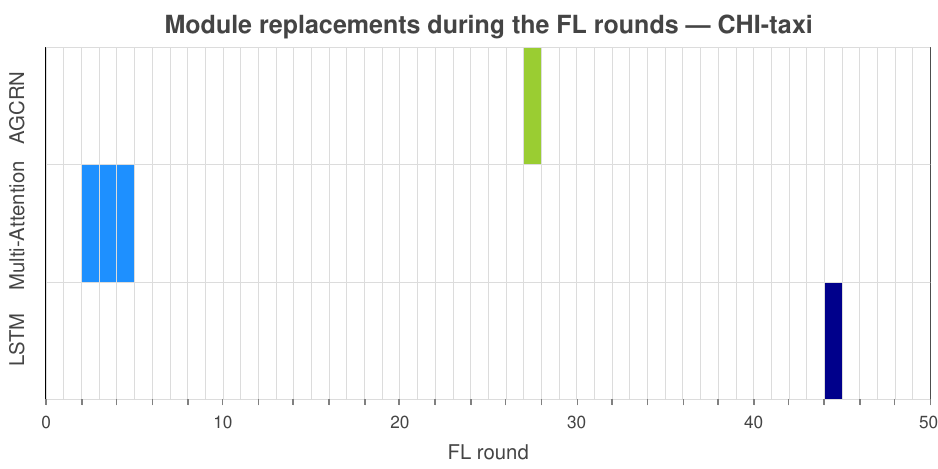}
    \caption{Module replacements map resulted by LSTM-DSTGCRN + FedAvg with CSV approach. The colored boxes indicate that at that FL round, the corresponding module was replaced by the aggregate one received from server.}\label{fig:replacement_map}
\end{figure}

To give an insight into the updates of the modules during the FL process, we visualize the replacements map resulted by LSTM-DSTGCRN + FedAvg + CSV approach in Figure \ref{fig:replacement_map}. We have trained the model for 50 FL rounds, and at each round, we record the module that was replaced by the aggregate one received from the server.
The colored boxes indicate that at that FL round, the corresponding module(s) was/were replaced by the aggregate one(s) received from server. The results show that the LSTM module is updated more frequently than the other modules, and the NYC-Bike and NYC-Taxi datasets have more frequent updates than the CHI-Taxi dataset. 
This suggests that these two clients are more active in the FL process, contributing more to the model's overall performance. Whereas, the CHI-Taxi dataset has fewer updates, indicating that the client's data may be less relevant (temporally and spatially) to the other clients' datasets.

{
Finally, to illustrate the complexity of the models and the communication overhead during the FL process, Table \ref{tab:payload} reports the number of parameters and the payload size for each local model in our experiments. As shown, the graph-based models require significantly larger payloads compared to GRU and LSTM. However, the payload sizes remain relatively small, suggesting that communication overhead is unlikely to become a major bottleneck during the FL process.
}
\begin{table*}[h]
    \centering
    \caption{Parameter and payload size of the approaches.}
    \label{tab:payload}
    \small
    {
    \begin{tabular}{lrrrr}
        \toprule
       Model & Number of parameters & Payload size \\
        \midrule
        GRU & 2753 & 14.81 KB \\
        LSTM & 3665 & 18.46 KB \\
        AGCRN \cite{BAIYao2020} & 119269 & 486.97 KB\\
        DSTGCRN \cite{GONG2024142581}& 121021 & 501.75 KB\\
        LSTM-DSTGCRN (ours) & 121933 & 505.39 KB\\
        \bottomrule
    \end{tabular}
    }
\end{table*}

\subsection{Ablation study of the LSTM-DSTGCRN model for transport demand datasets}
To underscore the critical importance of each component within the LSTM-DSTGCRN model, we conduct a ablation study for the transport demand datasets. 
This study systematically examines the contribution of individual components--namely, the LSTM, Multiahead Attention (MA) mechanism, and AGCRN, by selectively removing or altering them and observing the impact on model performance. Through this approach, we aim to quantify the significance of each component in capturing the complex spatiotemporal dependencies inherent in multimodal transport demand forecasting. 
By isolating the effects of these components, we can better understand their roles and the synergies they create within the overall model architecture, thereby validating the enhancements introduced in our modified approach.
\begin{table*}[h]
    \centering
    \caption{Ablation study of LSTM-DSTGCRN model on transport demand datasets}
    \label{tab:ablation}
    \begin{tabular}{lcccccc}
        \toprule
        \multirow{2}{*}{Model} & \multicolumn{2}{c}{NYC-Bike} & \multicolumn{2}{c}{NYC-Taxi} & \multicolumn{2}{c}{CHI-Taxi} \\
        \cmidrule(r){2-3} \cmidrule(r){4-5} \cmidrule(r){6-7}
        & MAE & RMSE & MAE & RMSE & MAE & RMSE \\
        \midrule
        Without LSTM & 1.9408 & 3.1432 & 10.1305 & \textbf{26.1436} & 3.3061 & 6.9760 \\
        Without MA & 1.9143 & 3.1485 & 9.9951 & 27.0137 & 2.9370 & 6.8889 \\
        Without LSTM and MA & 1.9661 & 3.1658 & 10.6862 & 26.7259 & 2.9259 & 6.9825 \\
        LSTM-DSTGCRN (our) & \textbf{1.8994} & \textbf{3.0450} & \textbf{9.8571} & 26.3968 & \textbf{2.6110} & \textbf{6.6846} \\
        \bottomrule
    \end{tabular}
\end{table*}

Table \ref{tab:ablation} presents the results of our ablation study on the three datasets. The metrics clearly demonstrate that each module within the LSTM-DSTGCRN model contributes significantly to its overall performance. 
The removal or alteration of any component, whether it be the LSTM, MA mechanism, leads to a noticeable decline in accuracy, as evidenced by increases in both MAE and RMSE. 
These findings underscore the critical role that each module plays in capturing the intricate spatiotemporal patterns and dependencies necessary for accurate transport demand forecasting.


\subsection{Comparison of local models and FL approaches on OD matrix datasets}
\label{sec:OD data}

Table \ref{tab:comparison_FL_years} provides a comparison of the performances of LSTM-DSTGCRN, GRU, and LSTM models with and without FL across two years (2021 and 2022) and two clients (Lyon PT and Orange Telecom) for OD matrix forecasting.

\begin{table*}[!h]
    \centering
    \caption{Comparison of the local learning and federated learning across years and clients}
    \label{tab:comparison_FL_years}
 {\scriptsize
    \begin{tabular}{lp{1cm}p{1.1cm}p{1.1cm}p{1.3cm}p{1cm}p{1.1cm}p{1.1cm}p{1.1cm}}
        \toprule
        \multirow{4}{*}{Model} & \multicolumn{4}{c}{2021} & \multicolumn{4}{c}{2022} \\
        \cmidrule(r){2-5} \cmidrule(r){6-9}
        & \multicolumn{2}{c}{Lyon PT} & \multicolumn{2}{c}{Orange Telecom} & \multicolumn{2}{c}{Lyon PT} & \multicolumn{2}{c}{Orange Telecom} \\
        \cmidrule(r){2-3} \cmidrule(r){4-5} \cmidrule(r){6-7} \cmidrule(r){8-9}
        & MAE & RMSE & MAE & RMSE & MAE & RMSE & MAE & RMSE \\
        \midrule
        GRU (locally) & 55.0299 & 134.6204 & 780.2223 & 1515.2890 & 47.5444 & 129.3204 & 927.7017 & 1645.8309 \\
        LSTM (locally) & 89.1214 & 278.9800 & 788.6166 & 1487.3965 & 86.2957 & 284.2191 & 925.4053 & 1617.0138 \\
LSTM-DSTGCRN & \textbf{42.3453} & \textbf{107.7701} & 722.3093 & 1137.5153 & 46.7632 & 158.5128 & 607.1526 & 939.9001 \\
   (locally) & \multicolumn{8}{c}{ }\\
        \midrule
        FedGRU & 96.8846 & 264.2408 & 773.3111 & 1525.6516 & 68.1231 & 160.2127 & 978.3316 & 1701.6402 \\
        FedLSTM & 92.0171 & 330.3783 & 780.8315 & 1423.6266 & 93.6214 & 283.8421 & 1061.4070 & 1711.0444 \\
        LSTM-DSTGCRN & 58.2769 & 127.7215 & \textbf{693.0593} & \textbf{1099.7407} & 44.0000 & 179.1663 & \textbf{579.1404} & \textbf{833.5996} \\
        + FedAvg & \multicolumn{8}{c}{ }\\
       LSTM-DSTGCRN & 46.1419 & 123.1041 & 737.1721 & 1193.0021 & \textbf{36.8111} & \textbf{89.4602} & 636.3386 & 917.6329 \\
  + FedAvg + CSV  & \multicolumn{8}{c}{ }\\
\multicolumn{8}{c}{ }\\
        \bottomrule
    \end{tabular}
}
\end{table*}

\begin{figure}[H]
    \centering
    \includegraphics[width=0.49\linewidth]{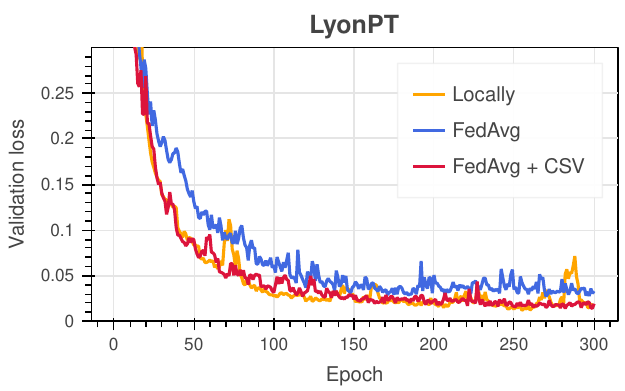}
    \includegraphics[width=0.49\linewidth]{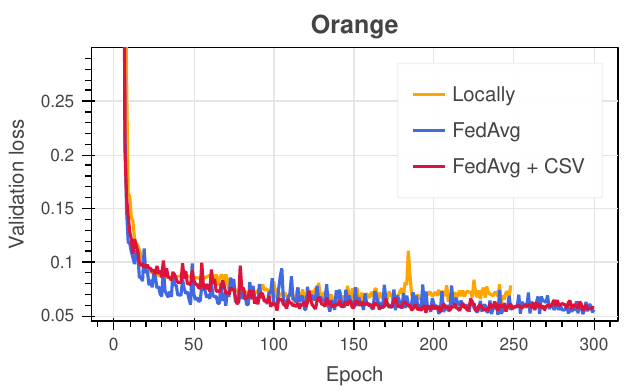}
    \caption{Validation loss of three learning approaches of LSTM-DSTGCRN model for OD data 2022.}
    \label{fig:val loss OD}
\end{figure}

\begin{figure}[!h]
    \centering
    \includegraphics[width=0.7\linewidth]{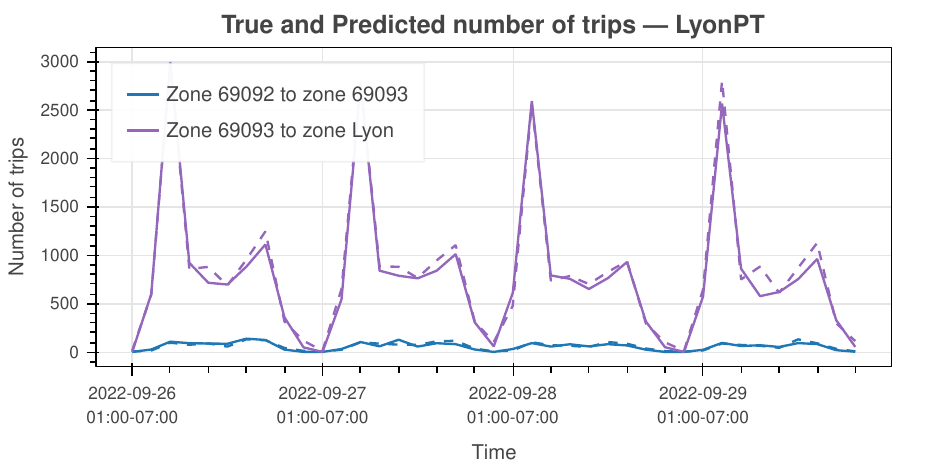}\\
    \includegraphics[width=0.7\linewidth]{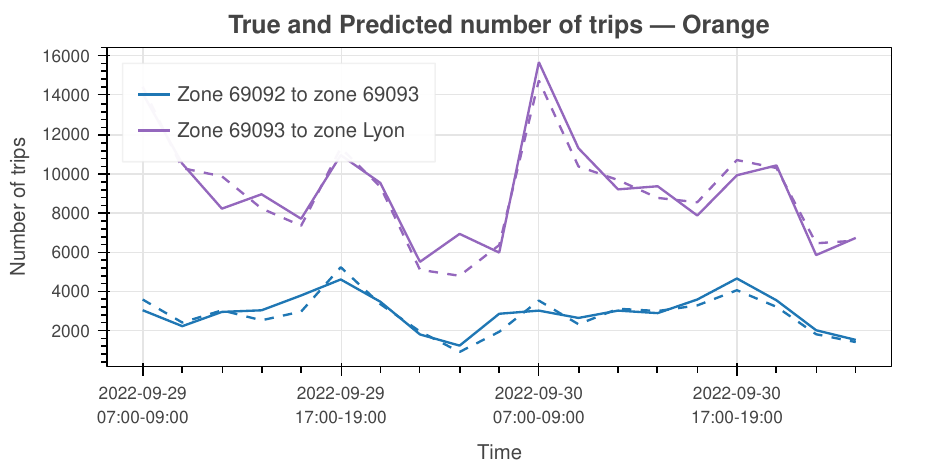}\\
    \caption{Forecasts given by LSTM-DSTGCRN + FedAvg with CSV at some random OD pairs. The dashed lines present the predicted values, the solid lines present the true values.}
    \label{fig:predictions OD}
\end{figure}

The comparison demonstrates the effectiveness of the proposed FL framework in enhancing forecasting accuracy compared to local learning methods. Notably, the LSTM-DSTGCRN model consistently achieves superior performance across all metrics, outperforming the GRU and LSTM models in both local and federated settings. 
This highlights its ability to effectively capture the spatiotemporal dependencies inherent in the OD data. Among the FL approaches, the integration of the CSV mechanism improves the model's performance on Lyon PT's data, but not on Orange Telecom's data. 
{We attribute this differential behavior to differences in data scale and industrial context. By construction, Orange's dataset is multimodal, encompassing not only public transport but also other transportation modes, and operates at a significantly higher scale than Lyon PT (as can be seen in Figure \ref{fig:predictions OD}). This scale difference may amplify the performance variations between methods, making subtle improvements less apparent in absolute terms. However, the CSV mechanism still provides valuable benefits for the Orange dataset. As demonstrated in Figure \ref{fig:val loss OD} (and discussed again later), the FedAvg + CSV approach (red line) closely tracks the FedAvg baseline (blue line) while exhibiting considerably greater stability throughout training, highlighting the beneficial effect of CSV in terms of prediction reliability even when absolute performance gains are minimal.}

The results in Table \ref{tab:comparison_FL_years}   emphasize  {again} the advantages of FL in leveraging collaborative training while preserving data privacy, particularly in 2022, where it demonstrates improvements for both clients. These findings underscore the practicality of the proposed FL framework for OD matrices forecasting.

Figure \ref{fig:val loss OD} shows the validation losses of the three approaches: LSTM-DSTGCRN with local learning, LSTM-DSTGCRN + FedAvg, and LSTM-DSTGCRN + FedAvg + CSV.  In particular, the LSTM-DSTGCRN with local learning (i.e., without federating) was trained with 300 epochs with early stopping technique (stop if the validation loss does not decrease for 10 consecutive epochs). 
The FL approaches were trained with 50 FL rounds and 6 local epochs per round. 
As we can see, for Lyon PT, learning with FedAvg + CSV results in comparable validation loss with local learning, even better in terms of reliability. 
For Orange, the FL approaches significantly improve the validation loss compared with local learning. 
We can see that for both clients, the CSV mechanism improved the stability of the losses, because, as designed, it filters out suboptimal updates, leading to more consistent and reliable model performance.

\begin{figure}[!h]
    \centering
    \includegraphics[width=0.7\linewidth]{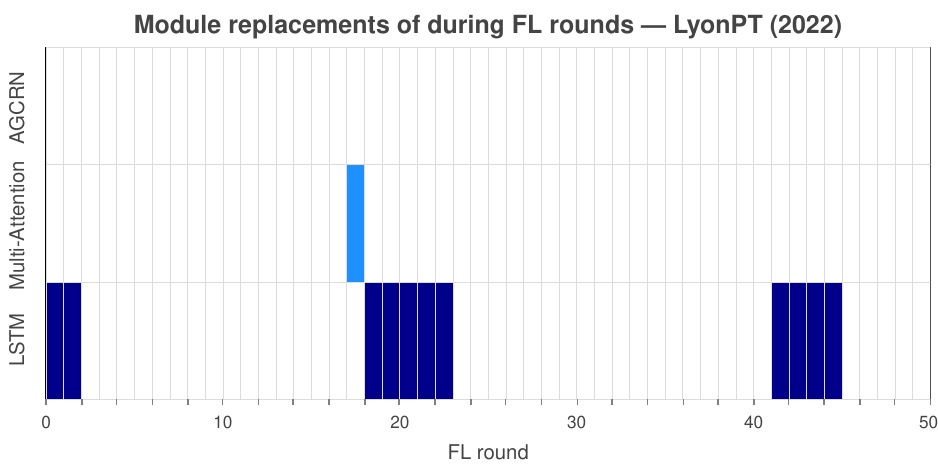}\\
    \includegraphics[width=0.7\linewidth]{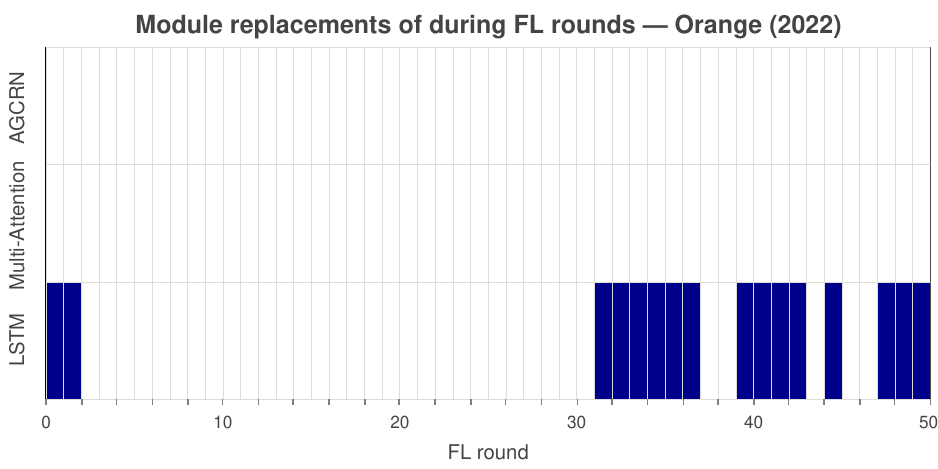}
    \caption{Module replacements map resulted by LSTM-DSTGCRN + FedAvg with CSV approach for OD data 2022. The colored boxes indicate that at that FL round, the corresponding module was replaced by the aggregate one received from server.}
    \label{fig:replacement_map_OD}
\end{figure}

Finally, for illustration, we show in Figure \ref{fig:replacement_map_OD} and Figure \ref{fig:predictions OD} the replacement map and the predictions resulted by LSTM-DSTGCRN + FedAvg + CSV approach for the OD data 2022.

 We can see that the LSTM module is updated more frequently than the other modules. This suggests that the clients learned from each other's temporal patterns, which are more relevant to the overall model's performance.

\subsection{The impacts of number of epochs and FL rounds on OD matrix data}
In this subsection, we {revisit the 2022 OD datasets to} investigate the {training  runtime and to analyze the} impacts of the number of epochs and FL rounds on the performance of the LSTM-DSTGCRN model with the proposed CSV mechanism and FegAvg scheme.  

{Table \ref{tab:runtime} summarizes the average runtime for both local and federated approaches corresponding to the experiments reported in Table \ref{tab:comparison_FL_years}. 
For local models, the runtime is averaged per epoch, enabling direct comparison of computational cost across architectures. 
For federated training, the runtime is averaged per round, encompassing local client updates, client-server communication, model aggregation, and the CSV selection process when applicable.
As expected, LSTM-DSTGCRN exhibits higher computational demands than the simpler GRU and LSTM baselines, requiring approximately 20 more time per epoch locally. This increase stems from the additional spatiotemporal graph convolution operations and attention mechanisms. 
The CSV mechanism introduces a modest overhead of approximately 5.5 seconds per round. This overhead reflects the additional computational cost of evaluating and selecting the validation-improving replacements of modules. Despite this increase, all runtimes remain within practical ranges, particularly considering the improved model stability and reliability that CSV provides.}
\begin{table*}[h!]
    \centering
    \caption{Average runtime across models on Lyon PT 2022 OD data.}
    \label{tab:runtime}
    \small
    {
    \begin{tabular}{lrr}
        \toprule
        Model & Average runtime  \\
        \midrule
        GRU (locally) & 0.232s/epoch  \\
        LSTM (locally) & 0.341s/epoch  \\
        LSTM-DSTGCRN (locally)  & 7.142s/epoch \\
        \midrule
        FedGRU & 1.512s/round \\
        FedLSTM & 2.0233s/round \\
        LSTM-DSTGCRN + FedAvg & 44.351s/round  \\
        LSTM-DSTGCRN + FedAvg + CSV  & 49.876s/round \\
        \bottomrule
    \end{tabular}
    }
\end{table*}

\begin{figure}[!h]
    \centering
    \includegraphics[width=0.49\linewidth]{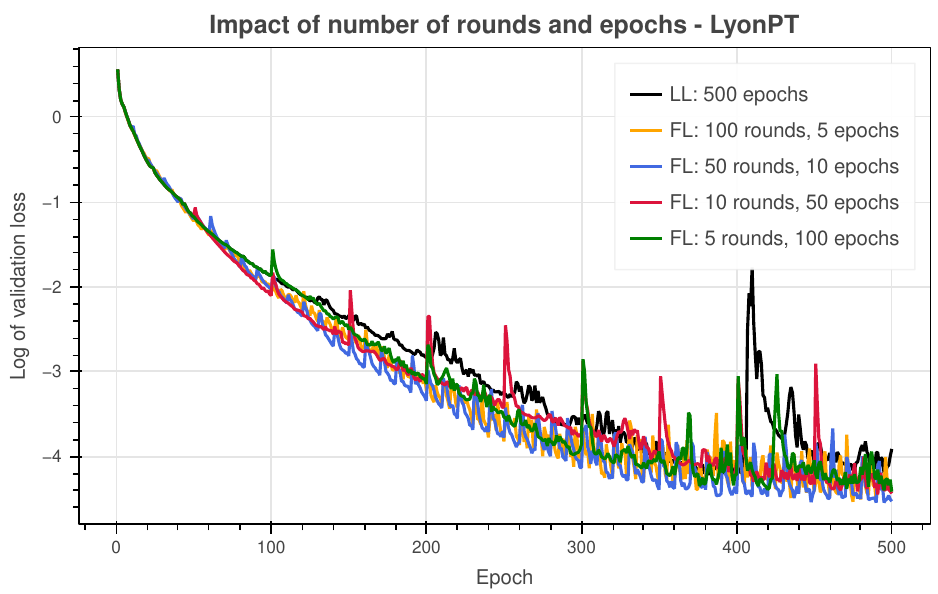}
    \includegraphics[width=0.49\linewidth]{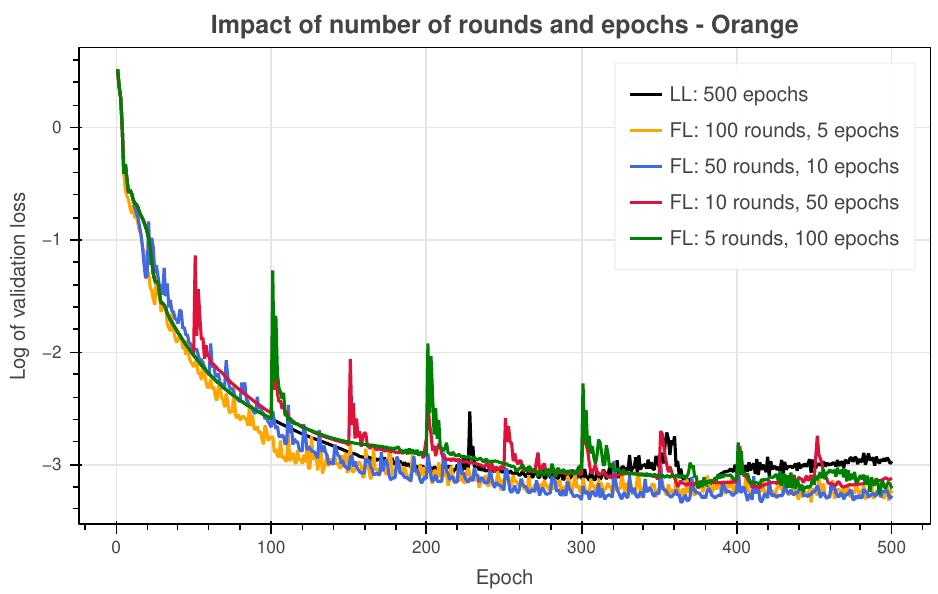}
    \caption{Impact of number of FL rounds and local epochs on learning prcoess of LSTM-DSTGCRN model for OD data 2022 {(LL stands for locally learning).}}
    \label{fig:rounds epochs}
\end{figure}

Figure \ref{fig:rounds epochs} shows validation losses (in log scale) of 5 different learning approaches of LSTM-DSTGCRN: Local learning with 500 epochs, LSTM-DSTGCRN + FedAvg + CSV with four combinations of FL rounds and local epochs (100-5, 50-10, 10-50, 5-100). 
The results  {show} that the learning process with 50 FL rounds and 10 local epochs per round {(blue line)} achieves the best performance, with the lowest and most stable validation loss. It is followed closely by the learning process with 100 FL rounds and 5 local epochs per round {(yellow line). These two approaches converge faster than the remaining three configurations. Particularly, the local learning with 500 epochs (black line) fails to provide stable convergence throughout the training process.}
This suggests that a balance between the number of FL rounds and local epochs is crucial for optimizing the model's performance. The learning process with fewer FL rounds tends to less effective. However, in general the FL processes give better performance than local learning, i.e., learning without collaboration.

\subsection{Limitations and potential improvements}

While our proposed framework shows improvements over existing methods, there are still some limitations and areas for future work:
\begin{itemize}
    \item \textbf{Cost of CSV step}: Although the CSV mechanism improves model performance, it may introduce additional computational costs due to the validation process, especially for clients wish to validate at layer level {or for deeper models. In these cases, the issue can be mitigated by ranking modules according to their gradient magnitudes (gradient-based importance) and validating only top-$k$ most important modules. Alternatively, an early stopping strategy can be applied, where the validation process is terminated once marginal improvements fall below a predefined threshold.}
    \item \textbf{Scalability}: Although the framework is designed to scale across multiple clients, similarly to the other FL framework, the communication overhead and computational complexity may increase with the number of clients. Future work could explore more efficient communication protocols and model compression techniques.
    \item \textbf{Data heterogeneity}: While the CSV mechanism addresses data heterogeneity to some extent, further improvements could be made by incorporating more sophisticated methods for handling diverse data distributions. {The combinations of FedProx \citep{li2020federatedoptimizationheterogeneousnetworks}, FedOpt \citep{reddi2021adaptivefederatedoptimization}, or FedDyn \citep{acar2021federatedlearningbaseddynamic} with CSV could be potential as these approaches are complementary, the former modify the update rule or adds regularization to stabilize training across non-IID clients, while the latter introduces an additional validation before accepting the aggregated parameter.}
    {
    \item \textbf{Security \& privacy-preserving}: CSV also acts as a lightweight local filter that reduces the propagation of low-quality or potentially corrupted updates, but it is not a full adversarial defense. A thorough security treatment would require pairing CSV with robust aggregation/defense mechanisms or privacy tools (e.g., secure aggregation).
    }
    \item \textbf{Extending to other domains}: The framework could be extended to other domains with similar spatiotemporal dynamics, such as energy consumption forecasting, environmental monitoring, {or communication} to validate its generalizability and effectiveness. {In energy consumption forecasting, we have graph structure with consumption points as nodes and power transmission lines as edges. We have also seasonal patterns and long-term dependencies as common as in transportation. However, for realistic applications in power grid management, incorporating physical power flow constraints and grid capacity limitations becomes essential to ensure feasibility of the predicted consumption patterns. For environmental monitoring problems, such as air quality forecasting or water quality forecasting, monitoring stations form nodes with atmospheric connections as edges. These data exhibit also seasonal patterns, long-term dependencies and different regions can mutually enhance their forecasting models through hidden pattern sharing via FL. Finally, in communication domain, traffic flows in mobile networks represent the movement of data packets between users and the network infrastructure that could be modeled as a graph structure where cellular towers or base stations serve as nodes and communication links or coverage overlaps form the edges.}
\end{itemize}

In conclusion, the proposed FL framework with LSTM-DSTGCRN and CSV presents a significant advancement in spatiotemporal forecasting, with promising results across multimodal transport demand and OD matrix datasets.

\section{Conclusion}
\label{sec:conclusion}
In this paper, we proposed an advanced FL framework to address the challenges of spatiotemporal data forecasting in diverse applications, including multimodal transport demand and OD matrix forecasting. By leveraging the enhanced LSTM-DSTGCRN model, our approach effectively captures the intricate spatial and temporal dependencies inherent in spatiotemporal systems while preserving data privacy through FL.

The LSTM-DSTGCRN model, with its integration of LSTM network, demonstrates superior predictive capabilities. It excels at modeling long-term temporal patterns and dynamic spatial interactions, making it particularly well-suited for applications where data is distributed across multiple entities, such as urban transportation systems and other decentralized spatiotemporal domains. 
The model's ability to adapt to the dynamic and heterogeneous nature of such systems ensures robust and accurate forecasting.

Our proposed FL framework extends existing methodologies by introducing a novel CSV mechanism. This mechanism allows each client to validate the aggregated parameters received from the server before integrating them into their local models. Experimental results show that the CSV mechanism significantly improves model robustness and accuracy by mitigating the effects of suboptimal updates. 
For instance, the FL approach with CSV achieved faster convergence and more stable training losses compared to traditional FL methods, as evidenced by experiments on both transport demand and OD datasets. 

In conclusion, the integration of the LSTM-DSTGCRN model within an FL framework with CSV represents a significant advancement in spatiotemporal data forecasting. This approach not only enhances model performance but also ensures data privacy and effectively handles data heterogeneity. Future research could explore further optimizations and extensions of this framework, including its application to other domains with similar spatiotemporal dynamics. These extensions would further validate the generalizability and impact of the proposed framework.

\section*{Acknowledgements}
\noindent This work was supported by the I-SITE FUTURE (reference ANR-16-IDEX-0003),  a program ``Investissement d'Avenir'' of France 2030 and for SystemX by the exploratory research facility
``EXPLO''.

\section*{References}
\bibliographystyle{elsarticle-harv}
{\small 
\bibliography{References-NeuComp}
}

\end{document}